\documentclass{article}

\usepackage[final]{neurips_2024}
\usepackage{enumitem}
\usepackage{algpseudocode}

\usepackage[utf8]{inputenc} % allow utf-8 input
\usepackage[T1]{fontenc}    % use 8-bit T1 fonts
\usepackage{hyperref}       % hyperlinks
\usepackage{url}            % simple URL typesetting
\usepackage{booktabs}       % professional-quality tables
\usepackage{amsfonts}       % blackboard math symbols
\usepackage{nicefrac}       % compact symbols for 1/2, etc.
\usepackage{microtype}      % microtypography
\usepackage{xcolor}         % colors
\usepackage{graphicx}
\usepackage{wrapfig}
\usepackage{amsmath}
\usepackage{natbib}
\usepackage{xspace}
\usepackage{amssymb}
\usepackage{float}
\usepackage[normalem]{ulem}
\usepackage{algorithm}
\usepackage{cleveref}
\usepackage{threeparttable}
\usepackage[normalem]{ulem}
\usepackage[most]{tcolorbox}
\usepackage{multirow}
\usepackage{makecell}
\useunder{\uline}{\ul}{}

\newcommand{\doubleunderline}[1]{%
    \underline{\underline{#1}}}

\newcommand{\eg}{\emph{e.g.},\xspace}
\newcommand{\ie}{\emph{i.e.},\xspace}
\newcommand{\etc}{etc.\xspace}
\newcommand\figref[1]{Figure~\ref{#1}}

\newcommand\tabref[1]{Table~\ref{#1}}

\newcommand\secref[1]{\S\ref{#1}}

\newcommand\appref[1]{Appendix~\ref{#1}}
\newcommand{\fakeparagraph}[1]
{\vspace{1mm}\noindent\textbf{#1}}

\definecolor{lightbluepurple}{RGB}{235, 240, 250}

\title{Concentrate Attention:\\Towards Domain-Generalizable Prompt \\Optimization for Language Models}
\author{Chengzhengxu Li\textsuperscript{$1$}, Xiaoming Liu\textsuperscript{$1,\ast$}, Zhaohan Zhang\textsuperscript{$2$}, Yichen Wang\textsuperscript{$3$}, \\ \textbf{Chen Liu\textsuperscript{$1$}, Yu Lan\textsuperscript{$1$}, Chao Shen\textsuperscript{$1$}}\\
  \textsuperscript{1}Faculty of Electronic and Information Engineering, Xi’an Jiaotong University\\
  \textsuperscript{2}Queen Mary University of London, London, UK  \quad 
  \textsuperscript{3}University of Chicago \\
  \textsuperscript{$\ast$} Corresponding author\\
  \texttt{\{czx.li, lcoder\}@stu.xjtu.edu.cn}\\
  \texttt{\{xm.liu, ylan2020, chaoshen\}@xjtu.edu.cn}\\
  \texttt{zhaohan.zhang@qmul.ac.uk} \quad \texttt{yichenzw@uchicago.edu} \\
}

\begin{document}

\maketitle

\vspace{-0.5cm}
\begin{abstract}
Recent advances in prompt optimization have notably enhanced the performance of pre-trained language models (PLMs) on downstream tasks.
% in few-shot natural language processing (NLP) scenarios.
However, the potential of optimized prompts on domain generalization has been under-explored.
% across unknown domains remains a significant challenge.
To explore the nature of prompt generalization on unknown domains, we conduct pilot experiments and find
that (\textit{i}) Prompts gaining more attention weight from PLMs’ deep layers are more generalizable and (\textit{ii}) Prompts with more stable attention distributions in PLMs’ deep layers are more generalizable.
Thus, we offer a fresh objective towards domain-generalizable prompts optimization named ``Concentration'', which represents the ``lookback'' attention from the current decoding token to the prompt tokens, to increase the attention strength on prompts and reduce the fluctuation of attention distribution.
We adapt this new objective to popular soft prompt and hard prompt optimization methods, respectively.
Extensive experiments demonstrate that our idea improves comparison prompt optimization methods by 1.42\% for soft prompt generalization and 2.16\% for hard prompt generalization in accuracy on the multi-source domain generalization setting, while maintaining satisfying in-domain performance.
The promising results validate the effectiveness of our proposed prompt optimization objective and provide key insights into domain-generalizable prompts.
Our codes are available at \url{https://github.com/czx-li/Concentrate-Attention}
\end{abstract}

\vspace{-0.3cm}
\section{Introduction}\label{Introduction}

Prompt optimization has emerged as a novel paradigm to effectively fine-tune pre-trained language models (PLMs), demonstrating impressive performance in natural language processing (NLP) tasks, especially under the few-shot setting \cite{schick2020exploiting,schick2020s,liu2023pre}.
Unlike traditional fine-tuning methods requiring training and saving entire model parameters \cite{devlin2018bert}, prompt optimization aims to explore well-performed prompts automatically in discrete or continuous space as a context for model input, which boosts model performance on downstream tasks.
The mainstream prompt optimization paradigms fall into two categories: \textit{hard prompt optimization }and \textit{soft prompt optimization}.
Hard prompt optimization relies on selecting well-performed prompts from a pre-constructed prompt set by filtering \cite{jiang2020can,haviv2021bertese,davison2019commonsense} or gradient-free optimization method \cite{li2024dialogue,sun2023query,prasad2022grips}.
Meanwhile, soft prompt optimization searches continuous embedding as prompts via gradient information guided by task-specific loss function \cite{vu2021spot, li2021prefix}.
%Traditional fine-tuning requires training and saving entire model parameters, which might be costly. 
% However, prompt optimization explores well-performed prompts automatically in discrete or continuous space as a context for model input, which boosts model performance on downstream tasks.

% Unlike traditional fine-tuning, which requires training and saving entire model parameters, prompt optimization aims to explore well-performed prompts automatically in discrete or continuous space as a context for model input and boost model performance on downstream tasks.
% guides PLMs to solve downstream tasks by combining inputs with special task-related text snippets which are optimized in discrete or continuous space through few training samples.
% % This method effectively harnesses the strong semantic capabilities of PLMs, transforming downstream tasks into mask-filling tasks through carefully crafted prompts. 
% Consequently, the model can address specific downstream tasks even in the few-shot setting.

% generates prompts in the form of readable natural language, which searches well-performed prompts in discrete space by editing tokens, designing evolutionary algorithms \citet{grips}, or training prompt selection agents.
% Meanwhile, soft prompt optimization searches continuous embedding as prompts via \todo{what} gradient information. 

However, while prompt optimization methods are becoming the mainstream of finetuning PLMs, the domain generalization ability of trained prompts still lacks exploration.
Previous works \cite{wu2022adversarial,zhao2022adpl,ge2023domain,guo2022improving} attempt to employ domain adaptation methods to address these challenges.
These works are based on the assumption of target domain availability.
They align the source domain and target domain by unsupervised feature learning.
The data reliance on these methods becomes a serious limitation for broader applications because models are frequently exposed to unknown domains.
Another branch to enhance the versatility of prompts is pre-training.
\citet{gu2021ppt} pre-trains prompts with 10 GB textual data.
\citet{vu2021spot} uses three tasks across eight datasets for pre-training to obtain transferable prompts.
As reported by \citet{liu2024stablept}, it requires 25-30 hours for pre-training prompts with Roberta-base on a single
NVIDIA A100.
The inefficiency and high computational cost remain a stumbling block for these methods to be widely used.
More importantly, the aforementioned methods are parameterized and not applicable to hard prompt optimization, showing low readability.
More studies refer to Appendix \ref{Related_Work}.

% Such work involves using a large amount of source domain data \citet{?} and unlabeled target domain data \citet{?} to participate in training.
% However, real-world applications require training-efficient prompts with few-shot setting that could generalize to data from unknown domains, which is beyond the capacity of these methods.
% Furthermore, current works \citet{?} attempt to pre-train prompts with enormous data from multiple domains or tasks, which are costly and not widely applicable to soft prompts. 
% This research gap impedes the efficiency of prompts deployment for each prompt could only be suitable for single domain but not universally applicable.

\begin{wrapfigure}[18]{r}{20em} 
    \begin{center}
    \vspace{-1.7em}
        \includegraphics[width=7.0cm]{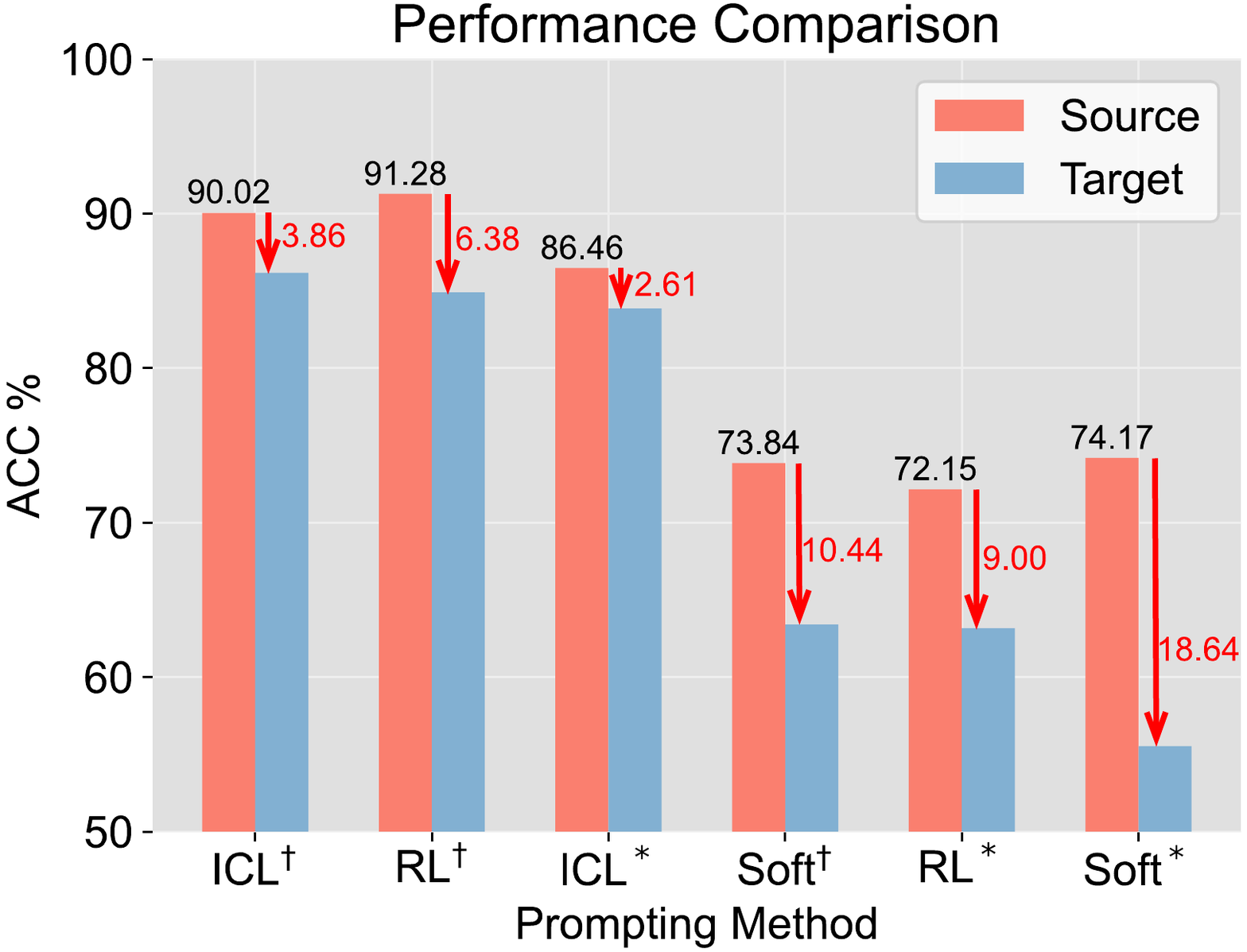}
        \setlength{\abovecaptionskip}{0.2cm} 
        \vspace{-1em}
        \caption{Domain generalization capabilities across various prompting methods (ICL \cite{brown2020language}, RL \cite{deng2022rlprompt}, Soft \cite{lester2021power}) in sentiment classification tasks.}% All prompts show performance degradation in the target domain.}
        \label{fig:acc}
    \end{center}
\end{wrapfigure}
Recognizing the problems mentioned above, 
we focus on improving the domain generalization ability of prompts with three constraints: (\textit{i}) do so with no knowledge about the target domain, 
(\textit{ii}) do so with little training cost,
(\textit{iii}) do so with easy adaptation on both soft prompt and hard prompt optimization.
To get started, we test the popular prompt optimization methods on cross-domain setting (\ie training prompts on one domain and testing them on out-of-distribution target domain\footnote{The $^{*}$ represents using CR \cite{hu2004mining} as the source domain and the  $^{\dagger}$ represents using SST-2 \cite{socher2013recursive} as the source domain. All results shown use MR \cite{pang2005seeing} as the target domain.}) and show the results in \figref{fig:acc}.
Interestingly, these optimized prompts exhibit (\textit{i}) great performance drop in general (by an average of 8.49\%) on target domain, validating the existence of research gap mentioned above, (\textit{ii}) different domain generalization ability in particular (Acc. drops by 2.61\% in best case and by 18.64\% in worst case), indicating the existence of distinct prompt ``nature'' that contributes to its generalizability.

Since prompts are functional in the model inference stage in which the model looks up contexts to generate new tokens through the attention mechanism, we probe the attention pattern on prompts during forward propagation with the question ``{\textit{what nature do well-generalized prompts have?}}'' %in the model inference stage.
and get the following findings ($\mathcal{F}$) via pilot experiments (\secref{pilot}):
%to analyze PLM's attention pattern when given different prompts and have the following discoveries:
% Specifically, in Section \ref{pilot}, we conduct pilot experiments to analyze PLM's attention pattern when given different prompts and have the following findings:
\begin{tcolorbox}[colback=lightbluepurple, colframe=white, rounded corners, 
                  left=3mm, right=3mm, top=1.5mm, bottom=1.5mm]

$\mathcal{F}_1$: Prompts gaining \textit{more attention weight} from PLMs' deep layers are more generalizable.

% PLMs' attention are \textit{more concentrated} on in deep layers are more generalizable.

$\mathcal{F}_2$: Prompts with \textit{more stable attention distributions} in PLMs' deep layers generalize better. %are more generalizable. 

\end{tcolorbox}

% \begin{itemize}[leftmargin=*]
% % \setlength{\leftmargin}{0pt}
% \item 
% \item 
% \end{itemize}
% (1) Prompts with better domain generalization capabilities attract more attention at deep layers of the PLMs; 
% (2) High-quality prompts exhibit more stable attention distributions at the deeper layers of the PLMs when combined with various inputs from target domains. 
% These discoveries indicate that prompts with satisfying domain generalization ability should be consistently "concentration-worthy" to PLMs.
Hence, we propose the idea of {\color[RGB]{0, 81, 186}{\textbf{\textit{Concentration}}}}, representing the capability of prompts to get the attention stably from PLMs. We suggest that the concentration indicates the domain generalization ability for prompts, which can be a forebode ahead of the downstream tests.

With the principle of concentration \secref{pilot}, we propose two algorithms that could piggyback upon popular prompt optimization methods for both hard and soft prompts to improve the domain generalization ability of prompts.
In the parameterized optimization process of soft prompt \secref{continuous}, where the loss function acts as objective, we introduce the concentration-reweighting loss.
It minimizes the attention weight on the original input sequence, so as to make the model concentrate on prompts stably for different inputs.
In the non-parameterized optimization process of hard prompt \secref{hard}, where the prompt set is first filtered and matched with different inputs by trained agents, we propose the concentration-oriented metric and reward.
They aim to filter out and match the input with concentration-worthy hard prompts.
% Further, we enhance both soft prompt and hard prompt optimization methods under the philosophy of concentration.
% Based on these discoveries, we reversely apply "concentration-worthy" as objective to search for universal prompts in both continuous and discrete space which could generalize well to unknown domain under few-shot setting with training inputs from multiple source domains. 
% For soft prompt optimization, we introduce the attention-reweighting loss, which minimizes the attention weight on the original input sequence, aiming to concentrate on prompts stably on different inputs.
%and keep this pattern  to different inputs.
% employs deep-layer attention of the PLMs as the loss for optimizing soft prompts without any other adjustments. 
% For hard prompt optimization, we propose a filter-match paradigm, which screens out concentrated prompts and then develops a multi-agent reinforcement learning (MARL) framework to achieve sample-level discrete prompt optimization. 
% We train the methods on inputs from multiple source domains and test them on an unknown domain under the few-shot setting.
Experiments show that our method respectively improves the target domain accuracy by 1.42\% and 2.16\% over the soft prompt and hard prompt optimized comparison methods, while maintaining in-domain capability.

% \todo{maybe  we should list the reason why we propose different methods for soft and hard prompts clearly in this section}

% propose two types of optimization methods suitable for soft and hard prompts, respectively.
% for the few-shot multi-source domain generalization settings
% \zzh{Based on the discovery above, we propose to exploit concentration-worthiness/attention-worthiness as an objective for prompt optimization, aiming to search for a universal template/prompt across multiple domains/tasks.
% Due to the different nature of continuous and discrete prompts, we design gradient-based and filter-and-matching-based? scheme for continuous and discrete prompts, respectively.} 
% These research outcomes provide new points and effective technical pathways for prompt optimization in few-shot multi-source domain generalization.
% \zzh{leave description of experiment to the very last}

%\czx{2024.5.1 To zhaohan and yichen

%1. There are too many Introduction parts and need to be streamlined;

%2. Whether the introduction in Figure 1 can be optimized and how to explain that RL(1) and RL(2), Soft(1) and Soft(2) are the same method but the source domain is different;

%3. Do Roberta-large and MR in Figure 1 need to be cited? I see that some articles do not add references here, but they are added in the subsequent introduction;
%}

%\zzh{Question: Do we need to compare the performance of in-domain prompt and out-of-domain prompt?}

\section{Preliminary}\label{Preliminary}
This section briefly introduces definitions of the Multi-source Few-shot Domain Generalization (MFDG) problem, which is the primary application scenario of our work.
%Subsequently we introduce our pilot experiments and applications in the following sections.

\fakeparagraph{MFDG Setting.} A text classification task, \eg sentiment classification, is defined as $\mathcal{T}:\mathcal{X} \to \mathcal{Y}$, where $\mathcal{Y}$ is the task’s label space and $\mathcal{X}$ is the feature space. 
We denote $M(X)$ to be the marginal distribution over $\mathcal{X}$, and $P(Y)$ to be the prior distribution over $\mathcal{Y}$.
The domain is then defined by $\mathcal{D}_{\mathcal{T}} = \left \{ \mathcal{X}, M(X), P(Y), P(Y|X) \right \}$.
Under the domain generalization setting, the source task is the same as the target task, \ie $\mathcal{T}_{s}$ equals to $\mathcal{T}_{t}$. 
But for the source domain $\mathcal{D}_{\mathcal{T}_{s}}$ and target domain $\mathcal{D}_{\mathcal{T}_{t}}$, at least one of the underlying probability distribution, \ie{} $M(X)$, $P(Y)$, or $P(Y|X)$, is different. 

In our MFDG problem, the training set is sampled from $N$ source domains $\mathcal{D}_{\text{train}} \sim \left \{\mathcal{D}_{\mathcal{T}_{s}}^{n} \right \}_{n=1}^{N}$ and the model is tested on an unknown target domain $\mathcal{D}_{\text{test}} \sim \mathcal{D}_{\mathcal{T}_{t}}$. Also, we follow \cite{perez2021true} to simulate the few-shot learning setting, which means $|\mathcal{D}_{\text{test}}| \gg |\mathcal{D}_{\text{train}}| $.

% \fakeparagraph{Few-shot.} Towards realistic scenarios, we propose a variant setting of the vanilla few-shot learning. We randomly select 16 inputs from each category $y \in \mathcal{Y}$ of each source domain dataset as the training set, while ensuring that the amount of training data in each source domains are equal and the sum is still much smaller than the target domain, which means $|\mathcal{D}_{\text{train}}|=N\times 16\times \left |\mathcal{Y}  \right | \ll |\mathcal{D}_{\text{test}}|$.

% \fakeparagraph{Prompting.} %Prompting is a promising solution to the MFDG problem.
% The method is to optimize prompt $z$, which includes a placeholder token \texttt{[MASK]}. And the final input $x$ for the downstream task is reformulated as $z \oplus x$.
% In the task, PLM is to predict which label token is most suitable to replace \texttt{[MASK]} and output it, \ie{} $\hat{y}=p_{\textrm{LM}}(z \oplus x)$.

\fakeparagraph{MFDG Objective.} Traditional prompting methods often rely on a crucial assumption that the training and testing sets come from the same underlying distribution $\mathcal{D}_{\text{train}}, \mathcal{D}_{\text{test}} \sim \mathcal{D}_{\mathcal{T}_{t}}$. 
In this context, the objective of prompting is to optimize high-quality prompt $z^{*}$ that maximizes the expected metric of the prediction on the target domain $\mathcal{D}_{\mathcal{T}_{t}}$:
\begin{align}\label{1}
z^{*} = \arg\max_{z} \mathbb{E}_{(x,y)\sim \mathcal{D}_{\mathcal{T}_{t}}} \left [ r(y, p_{\textrm{LM}}(z \oplus x)) \right ],
\end{align}
where $r$ is a function that evaluates the quality of the predicted answers when using the prompts $z$. 
For MFDG, the optimization objective is:
\begin{equation}\label{2}
z^{*} = \arg\max_{z} \mathbb{E}_{\mathcal{D}_{\mathcal{T}_{t}} \in \mathcal{G}} \left [ \mathbb{E}_{(x,y) \sim \mathcal{D}_{\mathcal{T}_{t}}} \left [ r(y, p_{\mathrm{LM}}(z \oplus x)) \right ] \right ],
\end{equation}
where $\mathcal{G}$ is the set of unknown target domains. 
In a nutshell, Eq. \ref{1} searches the prompts well-performed within the known domain, while Eq. \ref{2} explores the prompts that perform well across unknown domains.

\section{Concentration Benefits Generalization}\label{pilot}
In this section, we present pilot experiments to analyze the correlation between domain generalizability and attention concentration of prompts using RoBERTa-Large \cite{liu2019roberta} as the backbone. 
Appendix \ref{apdx:piolt} shows the specific form of prompts used in the pilot experiment.
From the effect of prompts in forward propagation, we analyze (\textit{i}) how much each prompt is concentrated by the LM, and (\textit{ii}) how stable the concentration is to formulate the correspondence.

%attention weights and their stability of PLMs in various prompts shown in Figure \ref{fig:acc} and discuss their effects on domain generalization performance. All experiments are based on the 354M-parameter PLM RoBERTa-large, and are conducted on the CR dataset as the target domain. The chosen prompts are based on key studies in few-shot prompt optimization. For detailed information on these prompts and the methods used, see Appendix A.

% \subsection{The Impact of Concentration Weights and Stability}\label{weights}

\fakeparagraph{Background.}
Attention mechanisms are widely studied for PLM interpretability \cite{wang2022paying,clark2019does,lin2019open,htut2019attention}. 
% It quantifies how the decision is made for currently decoded token by tracing back its attention weight $softmax(kq)$ on all previous tokens, where $k$ and $q$ are keys and quires in attention mechanisms.
As for prompt optimization, \citet{wang2023label} provide insights that label words in in-context learning aggregate most of the attention weights in deep layers of PLM, which majorly determine the final prediction. 
Inspired by this, we further explore the attention weight on the whole prompt sequence and its impact on prompt generalizability from a global perspective.

%\zzh{Given an $n$-token prompt $z=\left \{ e_{1}, e_{2},..., e_{n}^{mask} \right \} $ and a target domain input containing m tokens $x=\left \{ e_{1}, e_{2},..., e_{m} \right \} $, where $x \in \mathcal{X}_{\text{target}}$ and $e_{n}^{mask}$ represents the special [MASK] token. \zzh{variable overlap}
%The PLMs will generate the final prediction at the [MASK] token location. 
%The prompts and input are combined to form the final input text $z \oplus x=\left \{ e_{1}, e_{2},..., e_{n}^{mask}, e_{n+1}, e_{n+2},...,e_{n+m} \right \}$. This text will be input into PLMs, which generate corresponding predictions at the [MASK] token position.}
%\textcolor{yellow}{merge with section 2.} \yc{maybe a little bit too detailed here?}

\begin{figure}[h]
\centering
\includegraphics[width=13cm]{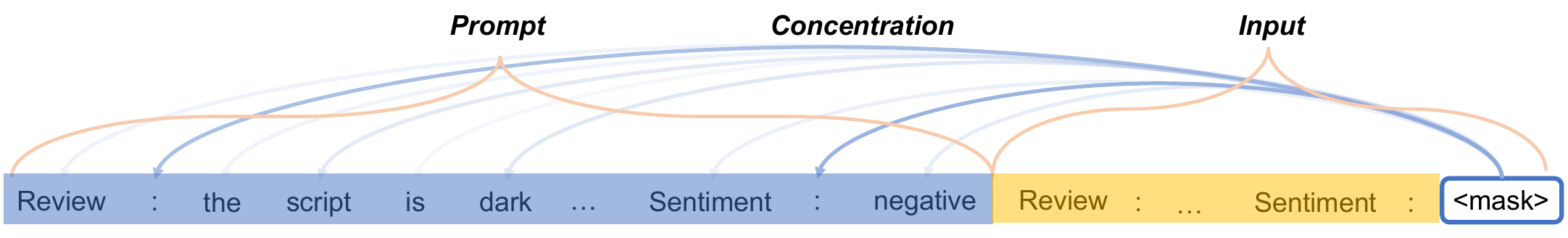} 
\caption{Illustration of Concentration. The tokens in the blue square are prompt, and those in yellow are input sequences. Concentration represents the model's attention on prompt tokens in forward pass when decoding <mask> token.}
\label{fig:concentration}
\end{figure}

\fakeparagraph{Definition 3.1.}
Let $z=(z_1, z_2,...,z_L)$ and $x =(e_1, e_2,...,e_T)$ be prompt and original input with $z,x \in S$, where $S$ is the set of all possible textual sequences over the vocabulary.
Let $f_{\theta_l}$ be the attention block\footnote{The attention block refers to key-query attention mechanism which is broadly used in transformers-based models.
The normalized, inner product of ``keys'' $k$ and ``queries'' $q$ is computed in the forward pass activations of attention block.} in layer $l$ of a PLM parameterized by $\theta_l$.
Then \textit{concentration} is a function $\text{Concentration}: S \rightarrow \mathbb{R}^+$ 
\begin{align}
    \text{Concentration}(z\oplus x;\theta_l) = \sum_{z_i \in z}f_{\theta_l}(z_{i} \oplus x).
\end{align}
Heuristicly, \textit{concentration} represents the ``lookback'' attention from current decoding token to prompt tokens, as shown in Figure \ref{fig:concentration}.

\fakeparagraph{Definition 3.2.}
Let $z=(z_1, z_2,...,z_L)$ and $x =(e_1, e_2,...,e_T)$ be prompt and original input with $z,x \in S$, where $S$ is the set of all possible textual sequences over the vocabulary.
Let $\mathcal{D} = (x_1, x_2,..., x_M)$ be the input dataset.
Let $f_{\theta_l}$ be the attention block in layer $l$ of a PLM.
Then \textit{concentration strength} is a function $\text{Strength}: \mathcal{D} \rightarrow \mathbb{R}^+$ 
\begin{align}
    \text{Strength}((z,\mathcal{D});\theta_l) = \frac{1}{|\mathcal{D}|} \sum_{x_i \in \mathcal{D}} \text{Concentration}(z\oplus x_i;\theta_l).
\end{align}
\textit{Concentration strength} represents the average concentration across the input dataset.

\fakeparagraph{Definition 3.3.}
Let $\mathcal{D} = {(x_1, x_2,..., x_M)}$ be the set of textual sequences sampled from target domain $\mathcal{\mathcal{D}}_{\mathcal{T}_{t}}$, where $x_i \in S$. Then the \textit{concentration fluctuation} is a function $\text{Fluctuation} : \mathcal{D} \rightarrow \mathbb{R}^+$ 
\begin{align}
    \text{Fluctuation}((z,\mathcal{D});\theta_l) = \sqrt{\frac{1}{|{\mathcal{D}}|}   \sum_{x_i \in \mathcal{D}} \left [\text{Concentration}(z\oplus x_i;\theta_l) )- \text{Strength}((z,\mathcal{D});\theta_l) \right ]^{2} } .
\end{align}
\textit{Concentration fluctuation} demonstrates the variance of concentration strength for different inputs.

\begin{figure}[h]
\centering
\includegraphics[width=14cm]{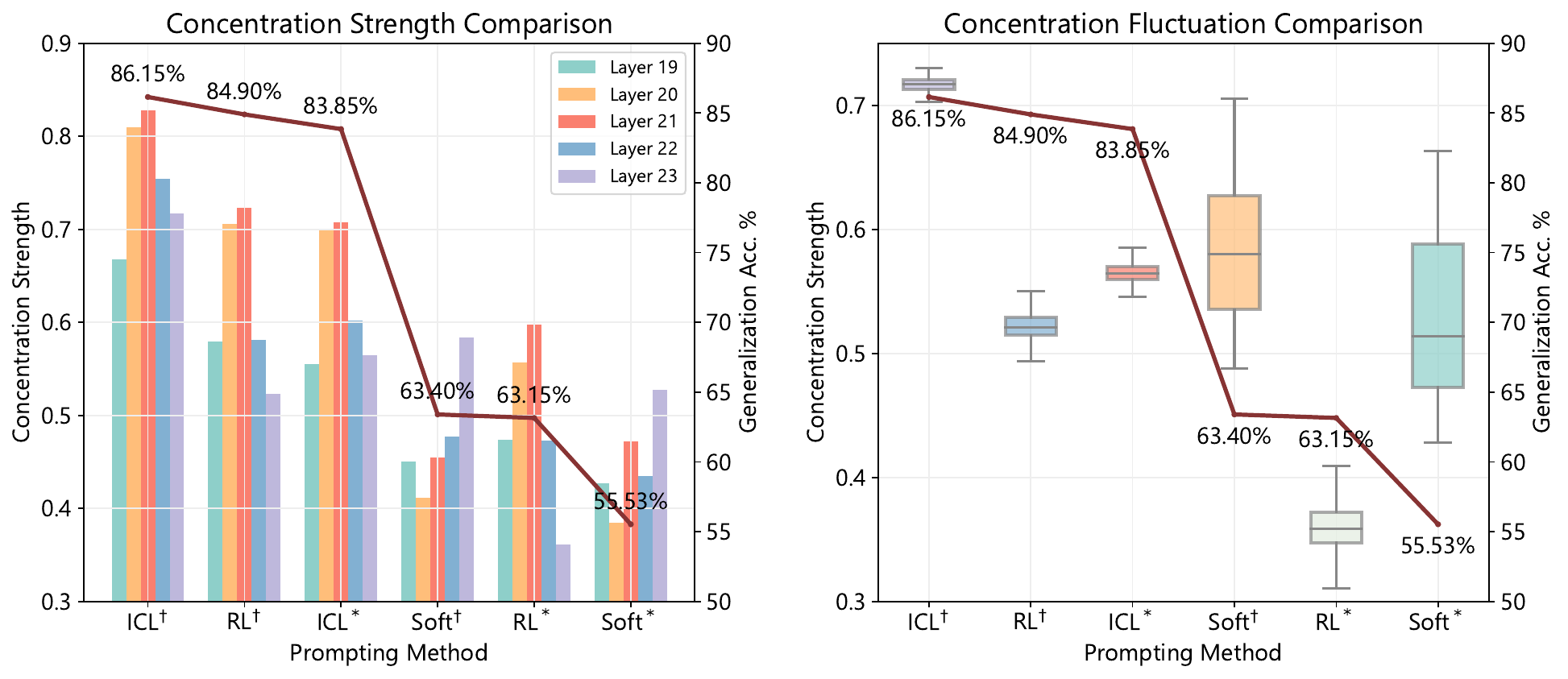} 
\caption{Left: concentration strength of various prompting methods in the last 5 layers (layers 19 to 23). Right: boxplots of the concentration strength in the last layer. Overall, prompts that exhibit good domain generalization gain higher concentration strength and lower concentration fluctuation. The concentration strength of each layer is shown in \appref{apdx:att-distribution}.}
\label{fig:rate}
\end{figure}

% \fakeparagraph{Results and Analysis.}
% As shown in \figref{fig:rate}, we exhibit the attention weights of the prompts in the last 5 layers of RoBERTa-large and their accuracy on the $\mathcal{D}_{t}$. 
Our pilot experiment unveils following insights: (\textit{i}) Prompts with larger Concentration Strength achieve better performance in domain generalization. 
For instance, Figure \ref{fig:rate}(left) shows that tokens of $\text{ICL}^{\dagger}$, the best-performed method, gain more than 0.8 of Concentration Strength at the 21st layer and over 0.7 at the 23rd layer.
(\textit{ii}) Prompts with lower Concentration Fluctuation tend to generalize to target domain better.
As shown in \figref{fig:rate}(right), $\text{Soft}^{\dagger}$ and $\text{ICL}^{*}$ are concentrated at a similar level, but $\text{ICL}^{*}$ generalizes better while its stability is better. 
(\textit{iii}) High Concentration Strength and low Concentration Fluctuation together contribute most to prompt generalizability.
The best-performed $\text{ICL}^{\dagger}$ has most Concentration Strength and lowest Concentration Fluctuation across all comparison prompts.
These discoveries inspire us to adjust the objective for soft prompt\secref{continuous} and hard prompt\secref{hard} optimization towards increasing Concentration Strength while decreasing Concentration Fluctuation.

\section{Concentrative Prompt Optimization}
\subsection{Concentrative Soft Prompt Optimization}\label{continuous}

\begin{wrapfigure}[11]{T}{21em} 
    \begin{center}
    \vspace{-0.7cm}
    \includegraphics[width=7.0cm]{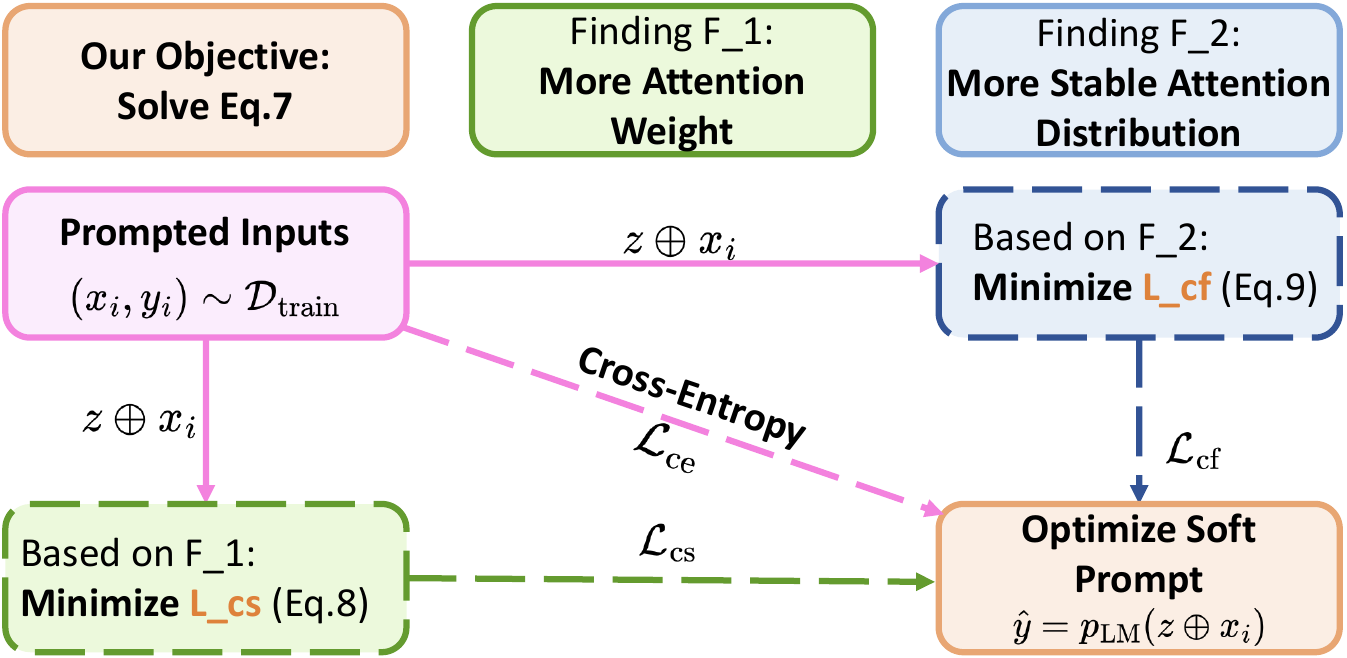}
    \setlength{\abovecaptionskip}{0.0cm}
        \caption{Framework for Soft Prompt Optimization.}
        \label{soft}
    \end{center}
\end{wrapfigure}

To devise soft prompt optimization with the guidance of concentration, we first visit the optimization objective of mainstream methods (\eg prompt tuning \cite{lester2021power}, prefix tuning \cite{li2021prefix}, p-tuning v2 \cite{liu2021p}).

These methods optimize follow log-likelihood objective given a trainable prompt $z$ and a fixed PLM parameterized by $\theta$ for the input $x$:
\begin{align} \label{eq:5}
    \max_{z}\text{log}P(y|(z\oplus x);\theta). 
\end{align} 

According to our findings in \secref{pilot}, domain-generalizable prompts should be high in concentration strength and low in concentration fluctuation.
Thus, we reformulate Eq. \ref{eq:5} to get the objective for domain-generalizable prompts:
\begin{align}\label{obj}
    \max_{z}(\text{log}P(y|(z\oplus x);\theta) + \text{Strength}((z,\mathcal{D}_{\text{train}});\theta)) \quad s.t. \  \min_{z} \text{Fluctuation}((z,\mathcal{D}_{\text{train}});\theta).
\end{align}

Towards the reformulated objective above, we propose the concentration-reweighting loss for soft prompt optimization methods.
The framework for soft prompt optimization is shown in Figure \ref{soft}.
First, we minimize the concentration strength on input $x$ to improve concentration strength on prompt $z$ by designing loss function $\mathcal{L}_{\text{cs}}$ as:
\begin{align}\label{eq:8}
   \mathcal{L}_{\text{cs}} = 1 - \text{Strength}((z,\mathcal{D}_{\text{train}});\theta).
\end{align}
%where $M$ is the batch size, $\alpha_{l}$ is the weight to balance the proportion of attention weights of each layer in $\mathcal{L}_{wgt}$ \xm{?}, and $\mathcal{L}$ is the number of layers in the PLM.

In addition, to reduce concentration fluctuation of prompts, we propose to use every token's concentration strength as hidden state feature of prompts, denoted as $\mathbb{C}_i = (c_1, c_2, ..., c_L)$ where $L$ is the length of prompts.
%$c_j = C_j(z \oplus x_i; \theta)$ and
We design a contrastive loss to cluster $\mathbb{C}$ with same label together to reduce concentration fluctuation:
\begin{align}\label{eq:9}
   \mathcal{L}_{\text{cf}}= \sum_{i=1}^{\left | \mathcal{D}_{\text{train}} \right |} \frac{-1}{P(i)} \sum_{p\in P(i)}\log\frac{\exp(sim(\mathbb{C}_{i},\mathbb{C}_{p})/\tau )}{ {\textstyle \sum_{j=1}^{\left | \mathcal{D}_{\text{train}} \right |}\mathbf{1}_{i\ne j}\exp(sim(\mathbb{C}_{i},\mathbb{C}_{j})/\tau )} } ,
\end{align}
where $P(j)$ represents the input with the same label as the $j$-th input in the dataset $\mathcal{D}_{\text{train}}$, $sim(.)$ is used to calculate the cosine similarity between feature embeddings,  $\mathbf{1}_{i\ne j}$ is an indicator function, \ie $\mathbf{1}_{i\ne j} \in {\left \{0, 1\right \}}=1$ if and only if $i\ne j$, and $\tau$ is a temperature parameter used to adjust the scale of the similarity score. 

Also, we utilize the cross-entropy classification loss $\mathcal{L}_{\text{ce}}$ \cite{mao2023cross}. %for improving the classification ability of soft prompts. 
The concentration-reweighting loss for soft prompt optimization is formulated as:
\begin{align}\label{eq:10}
    \mathcal{L_{\text{cr}}} = \lambda_{\text{ce}} \mathcal{L}_{\text{ce}} + \lambda_{\text{cs}} \mathcal{L}_{\text{cs}} + \lambda_{\text{cf}} \mathcal{L}_{\text{cf}},
\end{align}
where $\lambda_{\text{ce}}$, $\lambda_{\text{cs}}$ and $\lambda_{\text{cf}}$ weights different losses in training process. More details are in Appendix \ref{C}.

\subsection{Concentrative Hard Prompt Optimization}\label{hard}
\begin{wrapfigure}[11]{T}{21em} 
    \begin{center}
    \includegraphics[width=7.0cm]{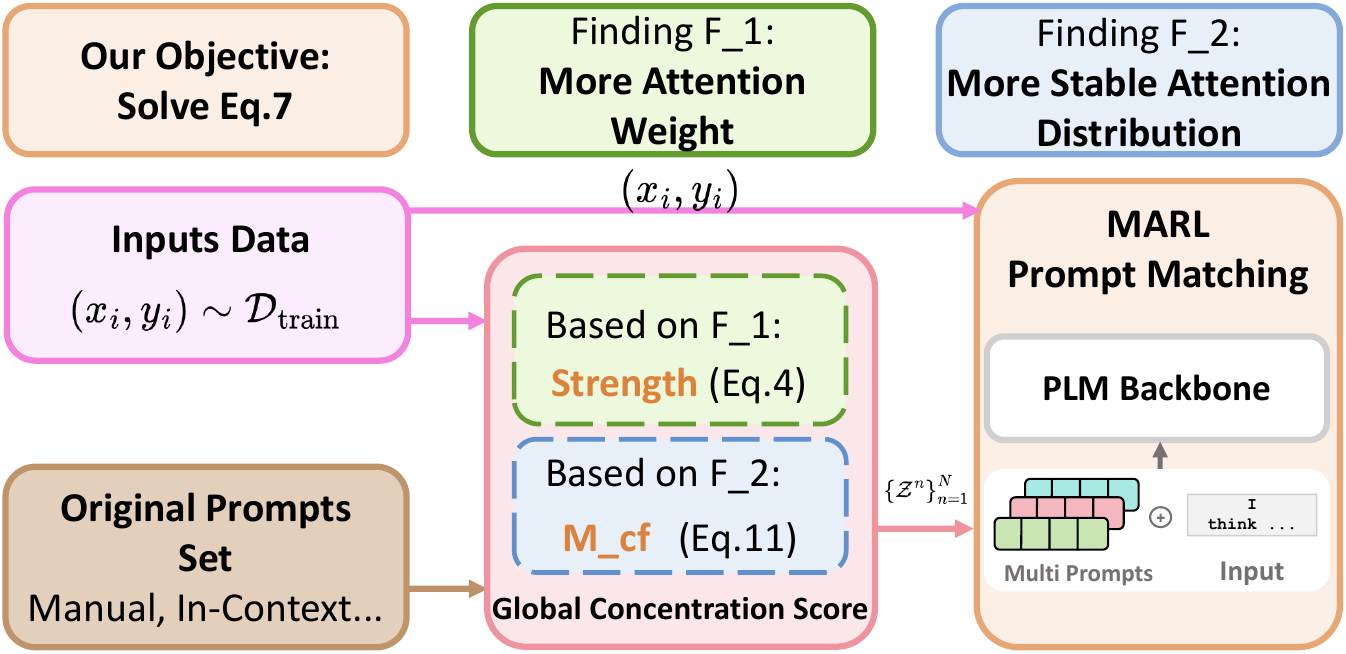}
    \setlength{\abovecaptionskip}{0.0cm}  
        \caption{Framework for Hard Prompt Optimization.}
        \label{fig:hard}
    \end{center}
\end{wrapfigure}
In contrast to soft prompt optimization, hard prompt optimization searches suitable prompt in discrete space in a non-parameterized fashion.
Previous hard prompt optimization searches can be divided into distribution-level \cite{prasad2022grips,deng2022rlprompt} and input-level \cite{li2024dialogue,lu2022dynamic}.
Although distribution-level prompt optimization can generally improve reasoning ability, motivated by the fact that no prompt is perfect for all inputs \cite{sun2023query}, we focus on improving the generalization ability of input-level optimization methods.
Generally, the mainstream of input-level optimization technique for hard prompts could be encapsulated as: \textbf{filter} (by \textbf{metric}) and \textbf{match} (by \textbf{RL agents}).
The findings of concentration could be applied to this optimization process by adjusting filter metric and agent reward.
We illustrate the framework for hard prompt optimization in Figure \ref{fig:hard}.

\fakeparagraph{Filter Metric.}
For previous filter metric only considering the overall accuracy on training set, we introduce a new metric called Global Concentration Score (GCS), which involves our ideas of concentration strength and concentration fluctuation.

Towards optimization objective Eq.\ref{obj}, we use concentration strength as first metric to filter out prompts which cannot get much concentration from model.
Metric for reducing concentration fluctuation could be regarded as minimizing Kullback-Leibler (KL) divergence between the concentration features $\mathbb{C}_i$ of input with same label and the average of $\mathbb{C}_i$ on whole inputs set $\mathcal{D}_{\text{train}}$:
\begin{align}
     M_{\textrm{cf}}(z, \mathcal{D}_{\text{train}}) = \sum_{y \in \mathcal{Y}}\sum_{i \in \mathcal{D}_{\text{train}}(y)} \text{KL}(\text{Softmax}(\mathbb{C}_{i}) \parallel \text{Softmax}(\mathbb{C}_{\text{avg}}^y)),
\end{align}

where $\mathcal{Y}$ is label space and $\mathcal{D}_{\text{train}}(y)$ is the input set labeled $y$ in data set $\mathcal{D}_{\text{train}}$. Also, we follow the setting of \cite{li2024dialogue} calculating the difference of the probability $p_{LM}$ that the $x_i$ is correctly labeled $y_{\text{ture}}$ and wrongly labeled as $y_{\text{false}}$ by a base PLM to improve the overall accuracy:
\begin{align}
    M_{\textrm{acc}}(z,\mathcal{D}_{\text{train}}) =\sum_{x_{i}\in{\mathcal{D}_{\text{train}}}} (p_{\textrm{LM}}(y_{\text{ture}}|z \oplus x_{i})-p_{\textrm{LM}}(y_{\text{false}}|z \oplus x_{i})).
\end{align}
Finally, we combine the above three metrics as one comprehensive metric,  \ie Global Concentration Score (GCS), to assess the quality of prompts:
\begin{align}\label{GCS}
\mathrm{GCS}(z,\mathcal{D}_{\text{train}})=\alpha _{\textrm{acc}}M_{\textrm{acc}}(z,\mathcal{D}_{\text{train}})+\alpha _{\textrm{cs}}\text{Strength}((z,\mathcal{D}_{\text{train}});\theta)+\alpha _{\textrm{cf}}M_{\textrm{cf}}(z, \mathcal{D}_{\text{train}}),
\end{align}
where $\alpha _{\textrm{acc}}$, $\alpha _{\textrm{cs}}$ and $\alpha _{\textrm{cf}}$ are the weights that balance accuracy, concentration strength, and concentration fluctuation, respectively.

\fakeparagraph{Prompt Matching.} 
% In previous studies, methods for optimizing prompts in reinforcement learning can be divided into two categories: generative and matching. Generative methods train an additional PLM through reinforcement learning to generate appropriate prompts for specific downstream tasks. The matching method performs sample-level optimization by assigning carefully designed prompts to different inputs to achieve optimal matching of discrete prompts. Although these methods have achieved certain results, they have some limitations. 
Previous methods mostly use a single RL agent to match appropriate prompts for each input \cite{li2024dialogue,lu2022dynamic,sun2023query}. Due to the large prompt space, the effective exploration of reinforcement learning agents is limited \citet{dulac2019challenges}. 
Furthermore, in the MFDG setting, inputs from different domains often have different state spaces, action spaces, and reward scales, then using a single agent often leads to the strategy converging to sub-optimality. 
To overcome these challenges, we redefine the discrete prompt matching problem in the MFDG setting as a multi-agent reinforcement learning (MARL) problem and propose a new matching algorithm.
% Due to the large prompt space, the effective exploration of reinforcement learning agents is limited. 
% Furthermore, in the MFDG setting, inputs from different domains often have different state spaces, action spaces, and reward scales, using a single agent often leads to the strategy converging to sub-optimality. To overcome these challenges, 
% we redefine the discrete prompt matching problem in the MFDG setting as a multi-agent reinforcement learning (MARL) problem and propose a new matching algorithm, shown in Algorithm 1.

We build our matching algorithm based on the Multi-Agent Proximal Policy Optimization (MAPPO) algorithm \cite{yu2022surprising}. Specifically, we configure one reinforcement learning (RL) agent for each source domain, collectively forming a multi-agent ensemble $\mathcal{N} = \{1, 2, \ldots, N\}$.
In order to effectively share learning experience in different domains, all agents share the same value network $v_{\phi}(.)$ while having independent strategy networks $\left\{\pi_{\omega_n}(.)\right\}_{n=1}^{N}$, where $\phi$ and $\omega_n$ are the learnable parameters.
Also, we define a set of prompts $\mathcal{Z}^{n}$ for each source domain, which serves as the action space for the corresponding RL agent. These prompts can come in various forms, including manual prompts \cite{bach2022promptsource}, original training inputs \cite{brown2020language,dong2022survey}, or LLM-generated \cite{li2024dialogue,lu2021fantastically}. Here, an action $a^{n}$ implies that agent $n$ selects a specific prompt $z^{n}$ from its designated prompt set $\mathcal{Z}^{n}$.

At each step $t$ of the training phase, given a state $s_{t}^n = \mathrm{PLM}(x_{t})$, which is the last hidden layer embedding of input $x_{t}$, the $n$-th agent selects an action $a_{t}^{n}$ by policy $\pi_{\omega_{n}}(a_{t}^{n}|s_{t}^n)$. This action corresponds to choosing prompt  $z_{t}^{n}$.
 We combine $x_t$ and $z_{t}^{n}$, feed them into the PLM for downstream tasks, and calculate the reward $r_{t}^{n}$. The agent’s parameters are then optimized based on $r_{t}^{n}$.

The rewards received by the RL agent are used as feedback to directly guide the optimization direction of the strategy. In this work, we aim to ensure that the prompts selected by the RL agent have good generalization capabilities. Therefore, we reuse $\text{Strength}(.;\theta)$ as a part of our reward function, specifically $r_{t}^{n}$ is defined as:
\begin{align}\label{reward}
    r_{t}^{n} = \alpha_{\textrm{acc}}M_{\textrm{acc}}(z_{t}^{n},\left\{x_{t}\right\})+\alpha_{\textrm{cs}}\text{Strength}(z,\left\{x_{t}\right\};\theta).
\end{align}

In the testing phase, we use an ensemble decision-making approach to apply the prompts. The prompts selected by each agent are input into the PLM to perform downstream tasks, and the results are combined. For a given input $x$ and its corresponding selected prompts $\left \{z^{n}\right\}_{n=1}^{N}$, the final prediction obtained by PLM for label $y$ can be expressed as:
\begin{align}\label{finalpred}
    P(y|x) = \mathrm{softmax}(\sum_{n = 1}^{N}p_{\mathrm{LM}}(y|x,z^{n})).
\end{align}

Our intention is to divide the action space of agents into smaller, more manageable subspaces and make it easier for agents to make the best decisions.
The detailed training and testing processes, along with specific agent settings, are presented in Appendix \ref{D}.

\section{Experiments}\label{Experiments}
To demonstrate the effectiveness of our findings for domain generalization, we conduct extensive experiments on tasks of sentiment classification and natural language inference (NLI).
We select the SST-2 \cite{socher2013recursive}, MR \cite{pang2005seeing}, and CR \cite{hu2004mining} datasets for sentiment classification, and the WNLI, QNLI, and RTE datasets from GLUE \cite{wang2018glue} for NLI tasks\footnote{For simplicity, these datasets are denoted by their initial letters (S, M, C, W, Q, and R respectively).}.
Each task involves designating one dataset as the target domain and the others as source domains. Detailed descriptions of the datasets and domain divisions are provided in Appendix \ref{dataset}.

We choose RoBERTa-large\cite{liu2019roberta} for all downstream tasks 
%because its model size is moderate (355M parameters) and suitable
for our hardware resources, and it has been widely used in previous prompt optimization works \cite{li2024dialogue,deng2022rlprompt,zhang2022tempera}.
Admittedly, at the time of writing this article, various efforts to optimize prompts have surfaced. 
However, our goal is not to build a better training method based on previous problems, but to pose a new problem, \eg learning prompts with strong domain generalization ability.
We therefore select three of the most well-known methods as baseline in the fields of soft prompt optimization and hard prompt optimization respectively. In addition, in order to more comprehensively demonstrate the performance of our method, we also select two distribution-level discrete prompt optimization methods as comparison methods.
The baseline methods and their implementations are described in Appendix \ref{apdx:baseline} and \ref{apdx:implement}.

\begin{table}[h]
\centering
\renewcommand\arraystretch{1.1}
\resizebox{\textwidth}{!}{
\begin{tabular}{cc|ccc|ccc}
\toprule \midrule
        &  \multicolumn{1}{c}{}    & \multicolumn{3}{c}{\textbf{Sentiment}}                                                           & \multicolumn{3}{c}{\textbf{NLI}}                                                              \\ \cmidrule(lr){3-5} \cmidrule(lr){6-8}
\textbf{\textit{Paradigms}} & \multicolumn{1}{c}{\textbf{\textit{Methods}}}       & S+M$\rightarrow$C & C+M$\rightarrow$S & \multicolumn{1}{c}{ S+C$\rightarrow$M }& Q+R$\rightarrow$W & W+R$\rightarrow$Q & Q+W$\rightarrow$R \\ \midrule
\multirow{4}{*}{\makecell[c]{Prompt \\ Tuning \\
\cite{lester2021power}}} &
Vanilla PT & 64.73$_{\text{3.82}}$                      & 65.51$_{\text{2.65}}$                       & 65.12$_{\text{3.85}}$                     & 41.20$_{\text{1.55}}$                   & {\ul 49.83$_{\text{1.47}}$}                & 49.66$_{\text{1.67}}$                     \\ 
&PT with $\mathcal{L}_{cs}$       & {\ul 65.83$_{\text{3.83}}$}                 & 66.38$_{\text{2.42}}$                      & 65.33$_{\text{2.37}}$                       & 41.53$_{\text{1.56}}$                      & 49.60$_{\text{2.37}}$                      & 49.43$_{\text{1.41}}$                     \\ 
&PT with $\mathcal{L}_{cf}$      & 65.09$_{\text{3.72}}$                       & {\ul 67.40$_{\text{2.43}}$}                 & {\ul 65.40$_{\text{2.33}}$}                 & {\ul 42.17$_{\text{2.03}}$}                & 49.22$_{\text{1.59}}$                     & {\ul 49.73$_{\text{1.31}}$}                \\ 
&PT with both     & \textbf{66.19}$_{\text{\textbf{3.69}}}$              & \textbf{69.54}$_{\text{\textbf{2.52}}}$              & \textbf{65.89}$_{\text{\textbf{2.32}}}$              & \textbf{42.48}$_{\text{\textbf{1.72}}}$             & \textbf{50.31}$_{\text{\textbf{1.33}}}$             & \textbf{50.42}$_{\text{\textbf{1.34}}}$             \\ \midrule

\multirow{4}{*}{\makecell[c]{Prefix \\ Tuning \\
\citet{li2021prefix}}} &
Vanilla Prefix & 65.91$_{\text{3.24}}$                      & 83.25$_{\text{0.41}}$                       & 75.51$_{\text{0.91}}$                     & 50.26$_{\text{0.31}}$                   & 51.88$_{\text{0.29}}$                & 50.02$_{\text{0.28}}$                     \\ 
&Prefix with $\mathcal{L}_{cs}$      &  66.23$_{\text{3.37}}$                 & {\ul 84.32$_{\text{0.48}}$}                      & 76.58$_{\text{0.82}}$                       & 50.69$_{\text{0.33}}$                      & 51.44$_{\text{0.28}}$                      & 49.77$_{\text{0.22}}$                     \\ 
&Prefix with  $\mathcal{L}_{cf}$     & {\ul 66.82$_{\text{3.19}}$}                       & 83.70$_{\text{0.39}}$                & {\ul 77.17$_{\text{0.75}}$}                 & {\ul 51.73$_{\text{0.32}}$}                & {\ul 52.12$_{\text{0.28}}$}                     & {\ul 50.73$_{\text{0.26}}$}                \\ 
&Prefix with  both     & \textbf{68.29}$_{\text{\textbf{2.97}}}$              & \textbf{85.07}$_{\text{\textbf{0.42}}}$              & \textbf{77.53}$_{\text{\textbf{0.43}}}$              & \textbf{52.05}$_{\text{\textbf{0.30}}}$             & \textbf{53.32}$_{\text{\textbf{0.25}}}$             & \textbf{51.26}$_{\text{\textbf{0.27}}}$             \\ \midrule

\multirow{4}{*}{\makecell[c]{P-Tuning \\ v2\\
\citet{liu2021p}}} &
Vanilla Pv2 & 65.92$_{\text{1.61}}$                      & 83.84$_{\text{1.69}}$                       & 75.89$_{\text{0.36}}$                     & 50.63$_{\text{0.31}}$                   & {\ul 52.76$_{\text{1.01}}$}                & {\ul 51.31$_{\text{1.37}}$}                     \\ 
&Pv2 with $\mathcal{L}_{cs}$      &  66.06$_{\text{1.77}}$                & 83.32$_{\text{1.59}}$                      & 75.07$_{\text{0.35}}$                       & {\ul 51.37$_{\text{0.37}}$}                      & 50.93$_{\text{0.92}}$                      & 50.20$_{\text{1.30}}$                     \\ 
&Pv2 with $\mathcal{L}_{cf}$      & {\ul 66.72$_{\text{1.62}}$}                       & {\ul 84.12$_{\text{1.51}}$}                 & {\ul 76.41$_{\text{0.33}}$}                 &  51.32$_{\text{0.38}}$                & 52.64$_{\text{1.04}}$                     & 51.28$_{\text{1.22}}$                \\ 
&Pv2 with both      & \textbf{67.07}$_{\text{\textbf{1.53}}}$              & \textbf{84.86}$_{\text{\textbf{1.42}}}$              & \textbf{77.26}$_{\text{\textbf{0.37}}}$              & \textbf{51.87}$_{\text{\textbf{0.28}}}$             & \textbf{53.83}$_{\text{\textbf{0.95}}}$             & \textbf{51.57}$_{\text{\textbf{1.16}}}$ \\ \midrule  \midrule

\makecell[c]{GrIPS
\\
\cite{prasad2022grips}} &
\makecell[c]{-}      & 80.07$_{\text{2.57}}$              & 84.28$_{\text{1.38}}$              & 85.19$_{\text{1.12}}$              & 54.37$_{\text{2.40}}$             & 52.77$_{\text{1.73}}$             & 53.52$_{\text{1.66}}$             \\ \midrule
\makecell[c]{RLPrompt
\\
\cite{deng2022rlprompt}}&
- & 86.05$_{\text{1.32}}$                      & 89.36$_{\text{0.91}}$                       & 85.95$_{\text{1.90}}$                     & 52.77$_{\text{2.82}}$                   &  53.82$_{\text{2.34}}$                &  54.63$_{\text{1.39}}$                     \\
\midrule
\multirow{4}{*}{\makecell[c]{Manual \\ Prompt \\
\citet{bach2022promptsource}}} &
Vanilla MP & 52.73$_{\text{4.43}}$                      & 55.81$_{\text{3.31}}$                       & 50.85$_{\text{1.58}}$                     & 41.70$_{\text{1.17}}$                   &  50.80$_{\text{0.84}}$                & 51.60$_{\text{1.50}}$                     \\ 
&MP with MARL       & {\ul 56.37$_{\text{1.18}}$}                 & {\ul58.42$_{\text{0.46}}$}                      & {\ul52.15$_{\text{0.49}}$}                       & {\ul44.27$_{\text{1.02}}$}                      & {\ul51.36$_{\text{0.84}}$}                      & {\ul52.18$_{\text{1.23}}$}                     \\ 
&MP with Metric      & 54.63$_{\text{2.12}}$                       &  57.84$_{\text{1.65}}$                 &  51.79$_{\text{1.75}}$                 &  42.86$_{\text{0.94
}}$                & 51.02$_{\text{0.68}}$                     &  52.03$_{\text{1.14}}$                \\ 
&MP with both     & \textbf{56.76}$_{\text{\textbf{0.40}}}$              & \textbf{59.44}$_{\text{\textbf{0.32}}}$              & \textbf{53.15}$_{\text{\textbf{0.35}}}$              & \textbf{45.05}$_{\text{\textbf{0.28}}}$             & \textbf{52.03}$_{\text{\textbf{0.25}}}$             & \textbf{52.46}$_{\text{\textbf{1.24}}}$             \\ \midrule

\multirow{4}{*}{\makecell[c]{In-Context \\ Demo \\
\cite{brown2020language}}} &
Vanilla IC & 84.33$_{\text{2.15}}$                     & 84.81$_{\text{1.39}}$                 & 80.21$_{\text{2.17}}$                 &  50.86$_{\text{1.28}}$                & 52.63$_{\text{0.94}}$                     & 58.04$_{\text{2.23}}$                     \\ 
&IC with MARL      &  {\ul85.33$_{\text{5.03}}$}                 & {\ul 87.02$_{\text{2.74}}$}                      & 82.14$_{\text{1.65}}$                       & {\ul52.82$_{\text{3.29}}$}                      & {\ul53.75$_{\text{1.32}}$}                      & {\ul59.87$_{\text{2.07}}$}                     \\ 
&IC with  Metric     &  84.70$_{\text{3.17}}$                       & 85.10$_{\text{2.12}}$                & {\ul 82.60$_{\text{4.91}}$}                 &  51.19$_{\text{4.80}}$                &  52.72$_{\text{4.31}}$                     &  59.46$_{\text{4.64}}$                \\ 
&IC with  both     & \textbf{87.29}$_{\text{\textbf{2.72}}}$              & \textbf{88.49}$_{\text{\textbf{1.52}}}$              & \textbf{83.52}$_{\text{\textbf{0.98}}}$              & \textbf{52.94}$_{\text{\textbf{1.59}}}$             & \textbf{54.24}$_{\text{\textbf{0.73}}}$             & \textbf{60.32}$_{\text{\textbf{1.20}}}$             \\ \midrule

\multirow{4}{*}{\makecell[c]{$\textsc{DP}_2\textsc{O}$ \\
\cite{li2024dialogue}}} &
Vanilla $\textsc{DP}_2\textsc{O}$ & 89.06$_{\text{0.76}}$                      & 90.75$_{\text{0.91}}$                       & 86.53$_{\text{0.80}}$                     & 54.84$_{\text{0.62}}$                   & 54.85$_{\text{0.37}}$                &  59.78$_{\text{0.79}}$                     \\ 
& $\textsc{DP}_2\textsc{O}$ with MARL      & {\ul87.36$_{\text{3.17}}$}              & {\ul91.60$_{\text{2.39}}$}              & 86.03$_{\text{4.03}}$              & {\ul54.71$_{\text{2.21}}$}             & 53.13$_{\text{1.97}}$             & {\ul60.62$_{\text{3.47}}$}                     \\ 
& $\textsc{DP}_2\textsc{O}$ with Metric      & 86.79$_{\text{1.32}}$              & 90.13$_{\text{1.07}}$              & {\ul86.60$_{\text{0.83}}$}              & 53.21$_{\text{1.16}}$             & {\ul54.02$_{\text{0.79}}$}             & 60.54$_{\text{1.47}}$                \\ 
& $\textsc{DP}_2\textsc{O}$ with both      & \textbf{89.63}$_{\text{\textbf{0.52}}}$                      & \textbf{92.87}$_{\text{\textbf{0.33}}}$                       & \textbf{87.85}$_{\text{\textbf{0.47}}}$                     & \textbf{56.42}$_{\text{\textbf{0.36}}}$                   &  \textbf{55.32}$_{\text{\textbf{0.33}}}$                &  \textbf{61.27}$_{\text{\textbf{0.81}}}$ \\ 

\midrule
\bottomrule
\end{tabular}
}
\caption{Performance comparison of text classification tasks in accuracy with MFDG setting. %Overall, our method improves the accuracy of the baseline methods in all cases. 
We use double horizontal lines to separate soft prompt optimization and hard prompt optimization methods.
``-'' denotes the distribution-level discrete prompt optimization methods which are not considered in our concentrative hard prompt optimization method, as stated in \secref{hard}.}

\label{PT}
\end{table}

\subsection{Out-of-domain Performance Comparison}
\fakeparagraph{Domain Generalization Result for Soft Prompt.}
As shown in Table \ref{PT},  the concentration strength loss $\mathcal{L}_\text{cs}$ and the concentration fluctuation loss $\mathcal{L}_\text{cf}$, in most experimental settings, enhance the domain generalization of three soft prompt optimization methods.
And, the combination of $\mathcal{L}_\text{cs}$ and $\mathcal{L}_\text{cf}$, \ie the concentration-reweighting loss $\mathcal{L_{\text{cr}}}$, further improves the domain generalization ability of soft prompts, achieving the best results in all experimental settings.
Specifically, $\mathcal{L_{\text{ar}}}$ (both) boosts the average accuracy of Prompt Tuning, Prefix Tuning, and P-Tuning v2 by 1.47\%, 1.78\%, and 1.02\%, respectively, highlighting its effectiveness in promoting the learning of domain-invariant properties in soft prompts. 
In addition, using only $\mathcal{L}_\text{cs}$ or $\mathcal{L}_\text{cf}$ alone may sometimes impair the performance of soft prompts, such as Prompt Tuning and P-Tuning v2 methods when QNLI data is used as the target domain. 
This indicates that concentration strength and concentration fluctuation are both indispensable for domain generalization ability of the prompts, and enhancing only one aspect may be harmful to the domain generalization performance of the prompts.

To more comprehensively illustrate the utility of the concentration-reweighting loss, we delve into Appendix \ref{C} for complete concentrative soft prompt optimization algorithm, extensive exploration on performance stability to prompt initialization and utility to decoder-only models of our method.
% Here, we examine its impacts on the initialization of soft prompt optimization, the outcomes within in-domain training, and the effects when employing GPT series models. 
Additionally, we provide quantitative analysis and visual representation to illustrate the impact of concentration-reweighting loss $\mathcal{L_{\text{cr}}}$ on soft prompts.

\fakeparagraph{Domain Generalization Result for Hard Prompt.}
% As for hard prompt, our proposed method has shown outstanding performance in domain generalization. 
% Specifically,
As shown in Table \ref{PT}, with the introduction of filtering metric and prompt matching framework, our approach effectively enhances the domain generalization capabilities of various existing methods. 
Among them, the improvements to the $\textsc{DP}_2\textsc{O}$ method achieved the best performance in all experimental setups. 
Compared to the original $\textsc{DP}_2\textsc{O}$, our method improve the average accuracy on sentiment classification and NLI tasks by 1.34\% and 0.85\% respectively. 
These results demonstrate the effectiveness of our proposed filtering metrics and surrogate rewards in selecting universal prompts from a pre-constructed set of prompts. 
Additionally, we find that compared to filtering metrics, the prompt matching framework brings a higher performance improvement to discrete prompts. 
This is because our reward function design adeptly guides the agent to match inputs with prompts that have strong cross-domain capabilities, even when faced with an unfiltered set of prompts. 
% Compared to generative method (\eg RLPrompt), our method uses filtered LLaMA-generated text as prompts. 
We also analyze our method from multiple aspects in Appendix \ref{D}. %including: complete concentrative hard prompt optimization algorithm, stability to hard prompt verbalizer selection, impact of the number of agents, and visualizations.

\fakeparagraph{Overall Comparison.}
In the MFDG setting, hard prompts generally outperform soft prompts. As illustrated in Table \ref{PT}, the best-performed hard prompt optimization method achieves a significant average accuracy of 73.88\%, compared to only 64.61\% for the best soft prompts. We hypnosis that hard prompts embed discrete tokens into the model input, providing precise guidance during testing and making it easy for PLMs to associate semantics fo input text sequence with the task. 
And soft prompts rely on indirectly influencing model inference by searching in continuous space with only the guidance of objective function, which might cause overfitting on source domain.

\subsection{In-domain Performance Comparison}\label{id}

\begin{table}[h]
\centering
\renewcommand\arraystretch{1.1}
\resizebox{\textwidth}{!}{
\begin{tabular}{cc|cccc|cccc|c}
\toprule
        &   \multicolumn{1}{c}{  }  & \multicolumn{4}{c}{Sentiment}                                                           & \multicolumn{4}{c}{NLI}                                                              \\ \cmidrule(lr){3-6} \cmidrule(lr){7-10} \cmidrule(lr){11-11}
\textbf{\textit{Paradigms}} & \multicolumn{1}{c}{\textbf{\textit{Methods}} }      & SST-2 & CR & MR & \multicolumn{1}{c}{Avg.} & WNLI & QNLI & RTE & \multicolumn{1}{c}{Avg.} & Avg Gap\\ \midrule
\multirow{2}{*}{\makecell[c]{Prompt \\ Tuning}} &
Vanilla PT & 73.84$_{\text{3.52}}$                     & 75.89$_{\text{1.72}}$                       & 74.17$_{\text{2.32}}$                     & 
74.63                     & 47.64$_{\text{1.02}}$                   & 49.71$_{\text{0.93}}$                & 54.73$_{\text{1.72}}$
&  50.69 & +6.66
\\ 
&PT with both     & 72.61$_{\text{2.72}}$              & 76.07$_{\text{2.24}}$              & 74.37$_{\text{2.12}}$   &74.35           & 46.79$_{\text{1.52}}$             & 49.50$_{\text{0.98}}$             & 54.21$_{\text{1.49}}$            & 50.17
&+4.71\\ \midrule

\multirow{2}{*}{\makecell[c]{Prefix \\ Tuning}} &
Vanilla Prefix & 87.39$_{\text{2.98}}$                      & 77.37$_{\text{0.79}}$                      & 82.65$_{\text{0.65}}$                     & 
82.47 &
55.88$_{\text{0.37}}$                   & 60.27$_{\text{0.44}}$                & 54.82$_{\text{0.31}}$        & 56.99
&+6.93 \\  
&Prefix with  both     & 87.29$_{\text{3.12}}$              & 76.73$_{\text{1.28}}$             & 83.32$_{\text{0.83}}$  & 82.45             & 56.18$_{\text{0.35}}$             & 59.74$_{\text{0.32}}$             & 55.38$_{\text{0.42}}$            &  57.10        &
+5.19\\ \midrule

\multirow{2}{*}{\makecell[c]{P-Tuning \\ v2}} &
Vanilla Pv2 & 86.71$_{\text{1.57}}$                     & 77.65$_{\text{1.49}}$                       & 82.27$_{\text{0.42}}$                     & 
82.21&
55.57$_{\text{0.73}}$                   & 60.73$_{\text{1.64}}$                & 55.16$_{\text{1.83}}$ &   57.15&                +6.29 \\ 
&Pv2 with both      & 87.03$_{\text{1.32}}$              & 77.71$_{\text{1.50}}$             & 82.05$_{\text{0.54}}$    &82.08          & 56.31$_{\text{0.69}}$             & 60.46$_{\text{1.37}}$             & 55.20$_{\text{1.69}}$ &57.32 &
+5.38\\ \midrule
\multirow{2}{*}{\makecell[c]{Manual \\ Prompt}} &Vanilla MP& 61.62$_{\text{3.42}}$                     & 57.75$_{\text{2.92}}$                       & 53.13$_{\text{2.33}}$                     & 
57.50&
44.27$_{\text{2.80}}$                   & 53.42$_{\text{0.98}}$                & 52.63$_{\text{0.60}}$&
50.11&  +3.22\\ 
&MP with both     & 61.33$_{\text{2.32}}$              & 56.07$_{\text{1.61}}$              & 53.47$_{\text{0.42}}$    &56.96          & 44.05$_{\text{0.89}}$              & 53.77$_{\text{1.35}}$              & 52.60$_{\text{0.39}}$              &  50.14&
+0.40\\ \midrule
\multirow{2}{*}{\makecell[c]{In-Context \\ Demo}} &Vanilla IC& 85.91$_{\text{1.42}}$                     & 85.57$_{\text{0.92}}$                       & 83.75$_{\text{1.39}}$                     & 85.08& 52.37$_{\text{1.45}}$                   & 53.42$_{\text{0.72}}$                & 59.73$_{\text{0.81}}$&
55.17&  +1.65\\ 
&IC with both     & 86.33$_{\text{1.34}}$              & 85.14$_{\text{0.87}}$              & 84.31$_{\text{2.12}}$   &85.26           & 52.25$_{\text{1.62}}$              & 52.96$_{\text{0.49}}$              & 59.36$_{\text{0.73}}$              &54.86&
+0.93\\ \midrule
\multirow{2}{*}{$\textsc{DP}_2\textsc{O}$} &Vanilla $\textsc{DP}_2\textsc{O}$ & 93.62$_{\text{0.72}}$                     & 90.76$_{\text{0.50}}$                       & 88.58$_{\text{0.91}}$                     & 90.99&
55.26$_{\text{1.02}}$                   & 55.13$_{\text{0.39}}$                & 61.07$_{\text{0.81}}$&  57.15&
+1.44\\ 
&$\textsc{DP}_2\textsc{O}$ with both     & 93.20$_{\text{0.81}}$              & 90.38$_{\text{0.47}}$              & 88.37$_{\text{2.12}}$   &90.65           & 56.47$_{\text{0.41}}$             & 55.42$_{\text{0.79}}$             & 61.29$_{\text{0.63}}$             &  57.73&
+0.37\\
\bottomrule
\end{tabular}
}
\caption{In-domain comparison. The last column shows the average gap between test performance on in-domain and out-of-domain data.}
\label{in-domain}
\end{table}

We also compare the in-domain performance between our proposed optimization objective and traditional training objective.
And we report not only model performance tested on in-domain dataset, but also the average gap between performance on in-domain and out-of-domain data. 
As shown in Table \ref{in-domain}, our method shows comparable accuracy with prompt optimization methods aiming only at maximizing log probability on correct label, demonstrating that taking concentration into consideration does not compromise on model performance on in-domain data.
Moreover, prompts optimized by concentration-driven objective shows better consistency when tested on both in-domain and out-of-domain data.
Especially, for hard prompt optimization which searches for suitable prompts in a limited discrete space, the average performance gap is less than 1\%, indicating our method always matches input sequence with proper prompts even if the prompts are not initially designed on target domain.

\subsection{Applicability to Larger Models and Other Tasks:}
We also attempt to extend our method to larger models and more complex tasks. We validate the effectiveness of our method on Llama-2-7b-chat \cite{touvron2023llama}, Vicuna-7b-v1.5 \cite{zheng2023judging}, and Alpaca-7b-wdiff \cite{alpaca} models for improving domain generalization ability of Prefix Tuning and In-Context Demo on question-answering tasks. We evaluate our method on ROC, SCT, and COPA datasets from the TRAM Benchmark \cite{wang2023tram} (referred as R, S, and C for simplicity), covering multiple choice question answering (MCQA) in reading comprehension and commonsense reasoning. The result is shown in Table \ref{tab:other}.

\begin{table}[h]
\centering
\renewcommand\arraystretch{1.2}
\resizebox{12cm}{!}{
\begin{tabular}{cc|cccc|c}
\toprule
  & \multicolumn{1}{c}{  }  &   \multicolumn{4}{c}{MCQA} \\ \cmidrule(lr){3-6} \cmidrule(lr){7-7}
\textbf{\textit{Models}} & \multicolumn{1}{c}{\textbf{\textit{Methods}}}       & S + C $\rightarrow$ R & C + R $\rightarrow$ S & R + S $\rightarrow$ C & \multicolumn{1}{c}{Avg.} & Acc Gap \\ \midrule
\multirow{4}{*}{\makecell[c]{Llama-2-7b-chat}} &
Vanilla Prefix &   62.32$_{\text{2.15}}$                      & 66.30$_{\text{2.30}}$                       & 73.15$_{\text{2.53}}$                        & 67.26        & ---                      \\ 
&Prefix with both     & 63.70$_{\text{1.96}}$              & 68.47$_{\text{0.97}}$              & 75.32$_{\text{1.09}}$ & 69.16  & +1.90 \\

&Vanilla IC     & 63.13$_{\text{1.25}}$              & 65.50$_{\text{1.98}}$              & 77.59$_{\text{1.14}}$  & 68.74 & ---\\

&IC with both     & 65.13$_{\text{1.03}}$              & 68.33$_{\text{2.13}}$              & 79.83$_{\text{0.88}}$  & 70.10 & +1.36 \\ \midrule

\multirow{4}{*}{\makecell[c]{Vicuna-7b-v1.5}} &
Vanilla Prefix &   67.72$_{\text{1.79}}$                      & 81.09$_{\text{2.17}}$                       & 88.97$_{\text{2.64}}$                                   & 79.26   & ---                 \\ 
&Prefix with both     & 68.75$_{\text{1.04}}$              & 83.93$_{\text{1.79}}$              & 89.76$_{\text{2.60}}$  & 80.81 & +1.55\\

&Vanilla IC     & 68.37$_{\text{2.24}}$              & 83.23$_{\text{4.12}}$              & 90.98$_{\text{1.99}}$  & 80.86 & --- \\

&IC with both     & 69.67$_{\text{1.58}}$              & 85.50$_{\text{5.06}}$              & 93.39$_{\text{1.23}}$  & 82.85 & +1.99 \\ \midrule

\multirow{4}{*}{\makecell[c]{Alpaca-7b-wdiff}} &
Vanilla Prefix &   61.52$_{\text{3.79}}$                      & 70.03$_{\text{2.88}}$                       & 87.91$_{\text{2.73}}$                             & 73.15            & ---              \\ 
&Prefix with both     & 63.89$_{\text{2.93}}$              & 72.15$_{\text{2.07}}$              & 89.58$_{\text{2.81}}$  & 75.21  & +2.06\\

&Vanilla IC     & 60.81$_{\text{1.14}}$              & 69.11$_{\text{2.46}}$              & 89.66$_{\text{2.37}}$  & 73.19 & ---\\

&IC with both     & 63.16$_{\text{1.56}}$              & 70.57$_{\text{1.95}}$              & 91.19$_{\text{2.00}}$  & 74.97 & +1.78\\
\bottomrule
\end{tabular}
}
\vspace{0.1cm}
\caption{Performance comparison of large models on MCQA task accuracy. The last column shows the average gap between test performance on vanilla method and our method.}
\label{tab:other}
\end{table}

Experimental results show that our method significantly improves the performance of large models on question-answering tasks across multiple domain generalization settings. For instance, for the Llama-7b model, our method improved the average accuracy of soft prompt generalization and hard prompt generalization comparisons by 1.90\% and 1.36\%, respectively; similar improvements were observed for Vicuna-7b and Alpaca-7b models, ranging from 1.55\% to 1.99\% and 2.06\% to 1.78\% respectively. 

Additionally, we would also like to discuss "\textit{why our method works well for large generative language models?}". In Appendix \ref{E}, we present the Concentration Strength Distribution of prompts using In-Context Demo across three 7B-sized language models (Llama, Vicuna, Alpaca) on three different tasks (SA, NLI, MCQA). We observe that all three LLMs exhibit stronger concentration strength in deeper layers compared to shallower layers when confront with prompts for different tasks. We find that this phenomenon occurs earlier in larger models (7B) compared to smaller models like Roberta-large. We speculate that this behavior is related to the alignment stage in pre-training of large models during Supervised Fine Tuning with a large number of prompts.

\section{Conclusion}
In this paper, we explore the nature of prompts with good domain generalization ability. 
By conducting experiments on model concentration on prompts and concentration pattern stability, we find that well-generalized prompt attract more attention weights at deeper layers of pre-trained language models (PLMs) and this pattern stably exists to different inputs. 
Inspired by these new findings, we propose optimization methods for soft prompt and hard prompt, respectively. 
For soft prompts, we design a concentration-reweighting loss to search for prompts with strong domain generalization ability in continuous space.
For hard prompts, we develop an attention-weighted filter-then-match framework.
This framework first apply a novel metric which takes model concentration and pattern stability into consideration to filter out low-quality prompts in candidate set.
Then a multi-agent reinforcement learning method is used to match each input with optimized hard prompts from each source domain.
Our extensive experiment on multiple datasets in different tasks demonstrates the superiority of our methods over existing comparison prompt optimization methods in terms of MFDG setting.

\section{Limitations}\label{Limitations}
In this study, we primarily focused on the performance of domain-generalizable prompt optimization. Despite this, our research still faces limitations in some practical application scenarios. Firstly, our pilot experiments only covered a limited variety of prompts. In future studies, we plan to extend to more diverse types of prompts. Secondly, the current research mainly focuses on the prompt domain generalization capabilities in a small-sample environment; next, we will conduct more comprehensive performance evaluations on complete datasets. Additionally, our current discrete prompt optimization method is primarily applicable at the input-level; in the future, we plan to explore its potential applications at the distribution-level. Finally, although our method is designed to enhance the performance of PLMs in classification tasks, these methods cannot be directly applied to open-ended generation tasks.

\vspace{0.1cm}
\section*{Acknowledgements}
We thank all the reviewers and the area chair for their helpful feedback, which aided us in greatly improving the paper.
This work is supported by National Natural Science Foundation of China (62272371, 62103323, U21B2018), Initiative Postdocs Supporting Program (BX20190275, BX20200270), China Postdoctoral Science Foundation (2019M663723, 2021M692565), Fundamental Research Funds for the Central Universities under grant (xzy012024144), and Shaanxi Province Key Industry Innovation Program (2021ZDLGY01-02).

% attention-based multi-agent reinforcement learning framework that outperforms existing state-of-the-art methods. These promising applications correspond to our findings and open up new research directions for future cue optimization research.
\bibliographystyle{abbrvnat}
\bibliography{neurips_2024}

\clearpage
\newpage
\appendix
\section*{Appendix}
\section{Related Work}\label{Related_Work}
\fakeparagraph{Prompting for Few-shot Learning.} Recent studies indicate that with pre-trained language models (PLMs) developing, prompt-based methods demonstrate significant competitiveness in downstream tasks with few-shot settings. 
For example, \cite{schick2020exploiting,schick2020s} propose a semi-supervised training method that converts the text classification task into a cloze task through word masking. 
Meanwhile, \cite{brown2020language,gao2020making,liu2023gpt} find that manual prompts can guide large machines to perform NLP tasks without any training. 
%With few-shot training examples, \citet{brown2020language,workrethinking,liu2021makes} achieve superior performance by inserting in-context demostrations.
\cite{vu2021spot, li2021prefix, an2022input, qian2022controllable} tune soft prompts using gradient descent with continuous embeddings instead of discrete prompts and achieve performance comparable to fine-tuning in few-shot setting.
Although these methods have demonstrated impressive performance, they often rely on a critical assumption, \ie the training and testing sets come from the same underlying distribution. 
Unfortunately, this assumption frequently does not hold in real-world scenarios.

\fakeparagraph{Domain Adaptation Prompting.} To address the out-of-domain challenges, many studies employ domain adaptation (DA) methods to acquire prompts that are effective in the target domain \cite{ge2023domain,guo2022improving,liu2022prompt,chen2024multi}. 
For example, \cite{wu2022adversarial} propose a novel domain adversarial training strategy to learn domain-invariant representations between each source domain and the target domain. \cite{zhao2022adpl} introduce three kinds of prompts learning task, source domain, and target domain features separately.
However, these methods still need the involvement of unlabeled target domain samples during training. 
In contrast to current research, our method expands the exploration of prompt optimization to the domain generalization problem where the target domain is entirely unknown during training.

\section{Experiment Setting Details}\label{A}
\subsection{Datasets} \label{dataset}
In \tabref{tab:4}, we provide details of the original datasets used in the main experiments, including type, domain, and label words, for tasks of sentiment analysis and  natural language inference (NLI).
\begin{table}[!h]
\centering
\renewcommand\arraystretch{1}
\tabcolsep=2pt
\begin{tabular}{ccccc} 
\toprule
Type &  Datasets & Domain &  Class  & Label words \\
\midrule
\multicolumn{1}{c}{\multirow{3}{*}{\begin{tabular}[c]{@{}c@{}}Sentiment\\ Analysis\end{tabular}}} & SST-2           &Movie Reviews &   2  & positive/negative                                                                   \\
\multicolumn{1}{c}{}                     & MR            & Movie Reviews                   & 2                    & positive/negative      \\
\multicolumn{1}{c}{}                     &CR              & Product    & 2       & positive/negative                                                                  \\ \midrule
\multicolumn{1}{c}{\multirow{3}{*}{NLI}}                     &RTE              & News   & 2     & Clearly/Yet                                                             \\
\multicolumn{1}{c}{}                     &QNLI               & Wikipedia  & 2       & Okay/Nonetheless                                                             \\
\multicolumn{1}{c}{}                     &WNLI               & Fiction Books   & 2          & Rather/Alas \\
\bottomrule
\end{tabular}
\caption{Datasets in the main experiments.}
\label{tab:4}
\end{table}

\tabref{tab:5} shows our specific division of the source and target domain under various settings of MFDG, as well as the sizes of the training and test set.
\begin{table}[!h]
\centering
\renewcommand\arraystretch{1}
\tabcolsep=2pt
\begin{tabular}{cccccc} 
\toprule
Type &  Setting & Source &  Target & |Train|/|Validation| & |Test|  \\
\midrule
\multicolumn{1}{c}{\multirow{3}{*}{\begin{tabular}[c]{@{}c@{}}Sentiment\\ Analysis\end{tabular}}} & S + M $\rightarrow$ C           &SST-2 \& MR &   CR&  64   &  2K                                                                  \\
\multicolumn{1}{c}{}                     & C + M $\rightarrow$ S            & CR \& MR       & SST-2            & 64                  & 1.8k        \\
\multicolumn{1}{c}{}                     &S + C $\rightarrow$ M              & SST-2 \& CR  & MR  & 64    & 2k                                                                     \\ \midrule
\multicolumn{1}{c}{\multirow{3}{*}{NLI}}                     &Q + R $\rightarrow$ W      & QNLI \& RTE        & WNLI   & 64   & 0.7k                                                               \\
\multicolumn{1}{c}{}                     &W + R $\rightarrow$ Q               & WNLI \& RTE        & QNLI & 64  & 5.4k                                                                 \\
\multicolumn{1}{c}{}                     &Q + W $\rightarrow$ R               & QNLI \& WNLI        & RTE   & 64   & 3K                                                                     \\
\bottomrule
\end{tabular}
\caption{MFDG setting for the main experiments.}
\label{tab:5}
\end{table}

\subsection{Baselines} \label{apdx:baseline}
We conduct extensive experiments, comparing 10 main competitors, including representative soft and hard prompting methods.

For the soft prompt optimization methods:
\noindent \textbf{Soft Prompt Tuning} \cite{lester2021power} replaces discrete prompt tokens with learnable embedding, and optimizes prompt through gradient information of PLMs. 
\noindent \textbf{Prefix Tuning} \citet{li2021prefix} reparametrizes networks for soft prompts and integrates and adjusts soft prompts at every layer of the PLM. 
\noindent \textbf{P-Tuning v2} \citet{liu2021p} is an improved version of Prefix Tuning, which has the option to reparameterize the network and use classification headers to adjust the soft prompts of each layer of PLM.

For the hard prompt optimization methods:
\noindent  \textbf{Manual Prompt} applies the prompt set designs of \citet{bach2022promptsource}, randomly combines the prompt with the input for downstream tasks.
\noindent  \textbf{In-Context Demo} \cite{brown2020language} randomly selects training data as examples to prompt PLMs to process subsequent inputs. 
\noindent \textbf{$\textsc{DP}_2\textsc{O}$} \cite{li2024dialogue} utilizes GPT-4 \cite{OpenAI2023GPT4TR} to generation a in-context prompt set and uses the reinforcement learning agent for prompt matching.
\noindent  \textbf{GrIPS} \cite{prasad2022grips} optimizes distribution-level hard prompts by editing on basic prompts, \ie{} substitution, deletion, and swapping, \etc 
\noindent \textbf{RLPrompt} \cite{deng2022rlprompt} uses reinforcement learning techniques to individually train partial parameters of PLMs to generate distribution-level discrete prompts for PLMs on downstream tasks.

\subsection{Implementation Details} \label{apdx:implement}
We provide experimental details for all baseline methods in the main experiment here. We choose RoBERTa-Large \cite{liu2019roberta} as our backbone model. 
We propose a variant setting of the vanilla few-shot learning \cite{perez2021true}.
For all tasks, we randomly select 32 samples from each source domain as the training set to simulate MFDG setting. We use the same approach to build the validation set and ensure that the number of labels in the training and validation sets is balanced. 
 For Soft Prompt Tuning, we replace the Manual Prompt tokens with five soft tokens in the same positions, and optimize them using AdamW \cite{loshchilov2017decoupled} optimizer with learning rate 2×10$^{-5}$ and batch size 32 for 300 epochs.
 For Prefix Tuning and P-Tuning v2, we apply the AdamW optimizer with a learning rate of 2×10$^{-4}$ and train for 100 epochs. 
 The mini batch size is 8 and prompt length is set as 10. 
 The setting of hard prompt optimization baselines (In-Context Demo, $\textsc{DP}_2\textsc{O}$, GrIPS and RLPrompt) follows \cite{li2024dialogue}.
 All experimental results are the average results of 10 different random seeds on a single NVIDIA A100 GPU.

\subsection{Training Details}
In this subsection, we provide additional details for reproducing our method. 
In prompt matching framework, each agent's policy network consists of two fully connected layers, $\omega_{n}^{1} \in \mathbb{R}^{1024\times 600}$ and $\omega_{n}^{2} \in \mathbb{R}^{600\times 15}$. 
The shared value network included three fully connected layers, sized $\phi^1 \in \mathbb{R}^{1024\times 600}$, $\phi^2 \in \mathbb{R}^{600\times 600}$ and $\phi^3 \in \mathbb{R}^{600\times 1}$. 
We use AdamW with eps of 0.00001 during training of 2000 epochs. The learning rate is 0.001, and mini-batch size is 32.
Also, in \tabref{tab:softweight} and \tabref{tab:hardweight}, we provide the balance weight settings of the soft prompt and hard prompt methods respectively.

\begin{table}[h]
\centering
\begin{minipage}[t]{0.48\textwidth}
\centering
\begin{tabular}{cccc} 
\toprule
  Method & $\lambda_{\text{ce}}$ &  $\lambda_{\text{cs}}$ & $\lambda_{\text{cf}}$  \\
\midrule
 PT with both &1 &     0.3   &  0.3                                                                  \\
Prefix with both            & 0.3                   & 0.5                  & 0.15        \\
Pv2 with both      & 0.5          & 0.5   & 0.15                                                             \\
\bottomrule
\end{tabular}
\caption{Weights for soft prompting methods.}
\label{tab:softweight}
\end{minipage}
\hfill
\begin{minipage}[t]{0.48\textwidth}
\centering
\begin{tabular}{cccc} 
\toprule
  Method & $\alpha_{\text{ce}}$ &  $\alpha_{\text{cs}}$ & $\alpha_{\text{cf}}$  \\
\midrule
 MP with both &10 &    7   &  7.5                                                                  \\
IC with both            & 7.5                   & 7.5                  & 0.15        \\
$\textsc{DP}_2\textsc{O}$ with both      & 10          & 6.5   & 6.5                                                               \\
\bottomrule
\end{tabular}
\caption{Weights for hard prompting methods.}
\label{tab:hardweight}
\end{minipage}
\end{table}

\vspace{-0.6cm}
\section{Pilot Experiments}\label{B}
\vspace{-0.1cm}
\begin{table*}[h]
\small
\centering
\renewcommand\arraystretch{1}
\begin{tabular}{>{\raggedleft\arraybackslash}p{0.10\textwidth} p{0.80\textwidth}}
\toprule
\textbf{$\text{ICL}^{\dagger}$:} & Review: the script is smart and dark - hallelujah for small favors. Sentiment: negative. Review: Good times to be found here if you love Rockabilly music. I'll definitely be be back here soon! Sentiment: positive. Review: <s> Sentiment: <mask>\\
\textbf{Method:} & In-Context
Demo \cite{brown2020language} \\
\midrule
\textbf{$\text{ICL}^{*}$:} & Review: it \'s just not very smart . Sentiment: negative. Review: extraordinary debut from josh koury. Sentiment: positive. Review: <s> Sentiment: <mask>\\
\textbf{Method:} & In-Context
Demo \cite{brown2020language}\\
\midrule
\textbf{$\text{RL}^{\dagger}$:} & <s> AgentMediaGradeOfficials Grade <mask>\\
\textbf{Method:} & RLPrompt \cite{deng2022rlprompt}\\
\midrule
\textbf{$\text{RL}^{*}$:} & <s> absoluteliterally absolute downright downright <mask>\\
\textbf{Method:} & RLPrompt \cite{deng2022rlprompt}\\
\midrule
\textbf{$\text{Soft}^{\dagger}$:} &<s> <soft> <soft> <soft> <soft> <soft> <mask>.\\
\textbf{Method:} & Soft Prompt Tuning  \cite{lester2021power}\\
\midrule
\textbf{$\text{Soft}^{*}$:} &<s> <soft> <soft> <soft> <soft> <soft> <mask>.\\
\textbf{Method:} & Soft Prompt Tuning  \cite{lester2021power}\\
\bottomrule
\end{tabular}
\caption{Prompt details of the pilot experiment in \secref{pilot}.}
\label{tab:6}
\end{table*}

\subsection{Pilot Details} \label{apdx:piolt}
To ensure a fair comparison between prompts of different lengths, we only select the top four tokens with the highest concentration strength in each prompt for experiment. We randomly select 1000 inputs from the target domain MR dataset for calculation. Results in \figref{fig:rate} reflect averages from ten random seeds. In \tabref{tab:6}, we show the specific methods and forms of the experimental comparison prompts in  \secref{pilot}.

\subsection{Attention Distribution Measurement}\label{apdx:att-distribution}

\figref{fig:istribution2} shows the distribution of concentration strength of various hints in each layer of the Robert-Large model in the pilot experiment. 
We can find that in PLMs, the concentration strength of almost all prompts is stronger in deep layers than in shallow layers, but there is a clear difference in their maximum values.
These findings prompt us to further investigate the properties of concentration strength.

\begin{figure}[h]
\centering
\includegraphics[width=6.8cm]{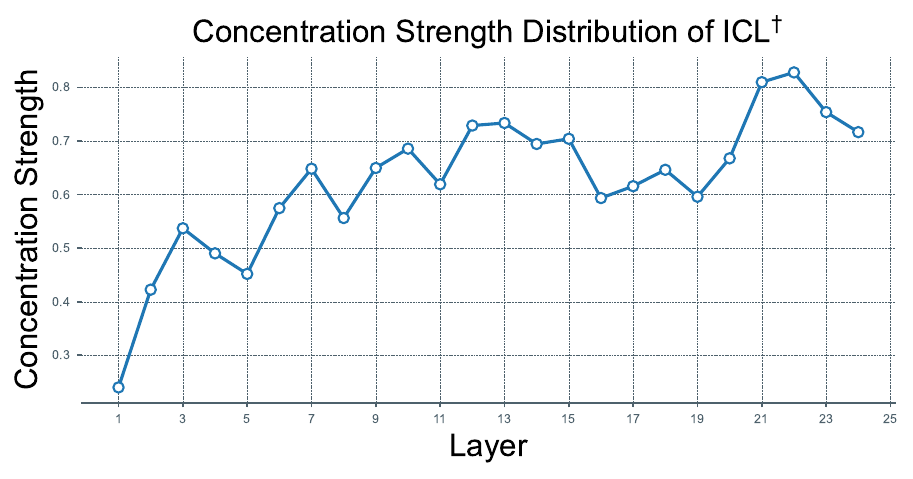} 
\includegraphics[width=6.8cm]{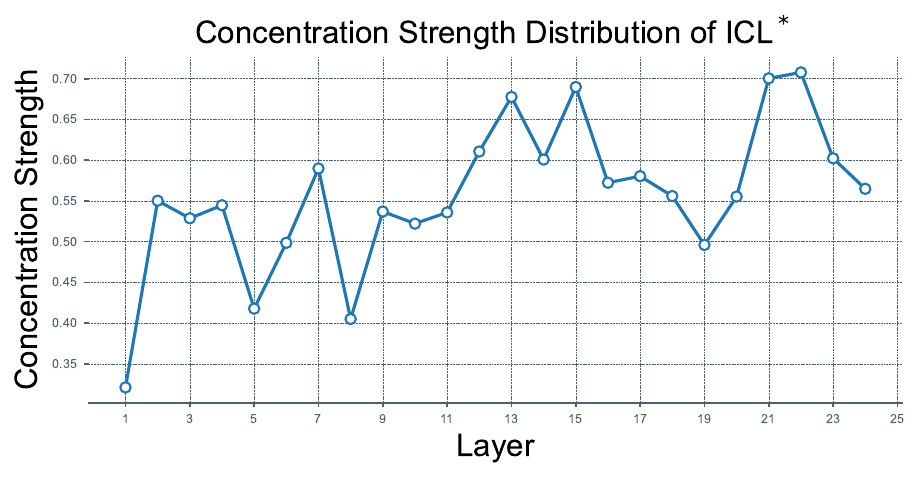}
\includegraphics[width=6.8cm]{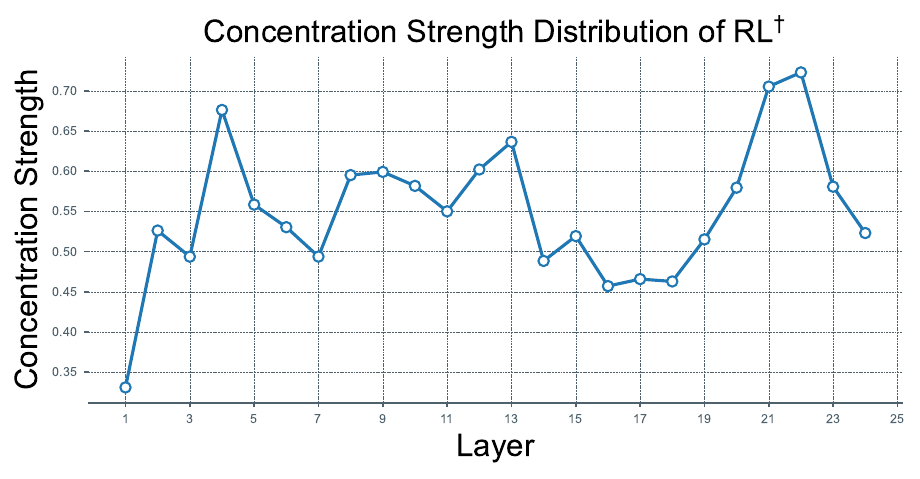}
\includegraphics[width=6.8cm]{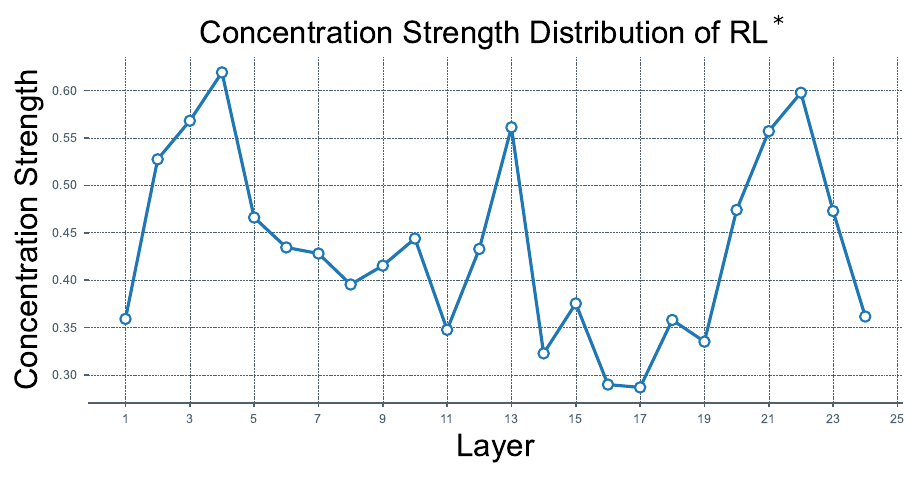}
\includegraphics[width=6.8cm]{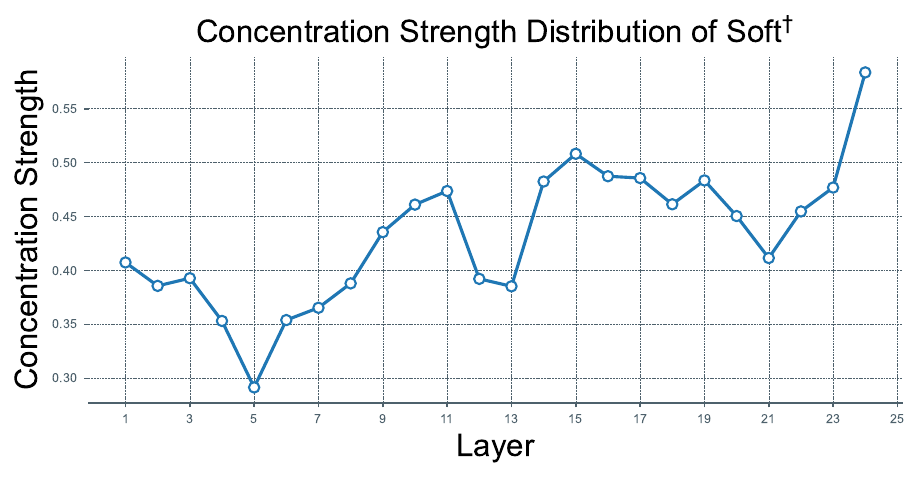}
\includegraphics[width=6.8cm]{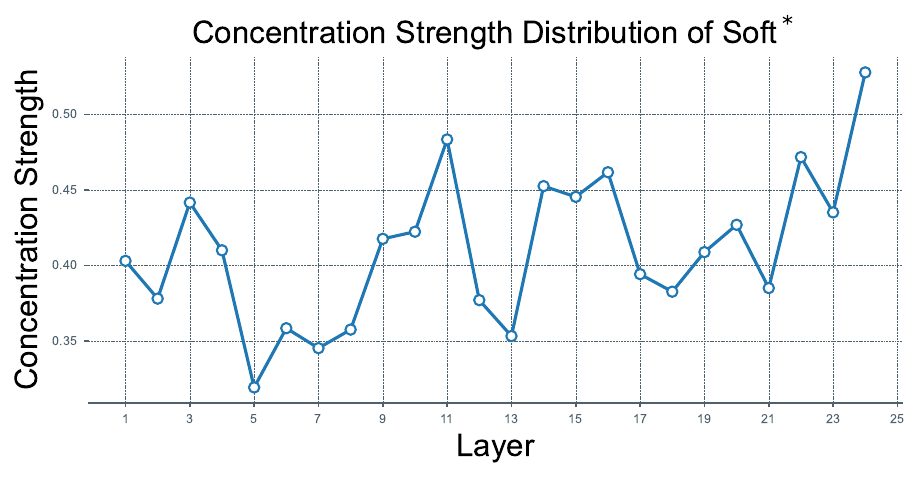}
\label{fig:istribution}
\caption{Distribution of concentration strength of various prompts in each layer of RoBERTa-Large.}
\label{fig:istribution2}
\end{figure}

\section{Details for Concentrative Soft Prompt Optimization}\label{C}

\subsection{Optimization Process} \label{apdx:soft-opt}
Algorithm \ref{algorithm:1} shows the detailed process of concentrative soft prompt optimization in \secref{continuous}. 
It also reveals that our method can be widely applied to different soft prompt optimization methods to improve their domain generalization capabilities.
\begin{algorithm}
\caption{Concentrative Soft Prompt Optimization}
\label{algorithm:1}
\begin{algorithmic}[1]
\State \textbf{Input:} fixed PLM parameterized by $\theta$, training dataset $\mathcal{D}_{\text{train}}$, learning rate $\eta$, loss weight $\lambda_{\text{cs}}$, $\lambda_{\text{cf}}$ and $\lambda_{\text{ce}}$
\State Initialize soft prompts $z_{\text{soft}}$ with random values
\While{not converged}
    \For{each input $(x_i, y_i)$ in $\mathcal{D}_{\text{train}}$}
        \State Construct input sequence $ (z \oplus x_{i})$ 
        \State Get final prediction $P(\hat{y_{i}}|(z \oplus x_{i});\theta)$
        \State Compute loss $\mathcal{L}_{\text{cs}}$ by Eq. \eqref{eq:8}
        \State Compute loss $\mathcal{L}_{\text{cf}}$ by Eq. \eqref{eq:9}
        \State Compute loss $\mathcal{L}_{\text{ce}}$ by cross-entropy classification loss \cite{mao2023cross}
        \State Compute final loss by Eq. \eqref{eq:10}
        \State Compute gradients $\nabla z_{\text{soft}} = \frac{\partial L}{\partial z_{\text{soft}}}$
        \State Update prompts $z_{\text{soft}} \leftarrow z_{\text{soft}} - \eta \nabla z_{\text{soft}}$
    \EndFor
\EndWhile
\State \textbf{Output:} Trained domain-generalizable soft prompt $z_{\text{soft}}$
\end{algorithmic}
\end{algorithm}

\subsection{Stability to Soft Prompt Initialization}\label{apdx:stable_soft_initial}
We adopt five different soft prompt initialization strategies \cite{gu2021ppt} to test the stability of our method. 
``Random'' indicates that we randomly initialize the embedding of soft prompt. 
``Label'' indicates that we use the embeddings of the label words. 
``Vocab'' indicates that we randomly sample words from the vocabulary. 
``Top-1k'' indicates that we randomly sample words from the most frequent 1000 words in the pre-training corpus. 
``Task'' indicates that we randomly sample words from the downstream data.

As shown in \tabref{tab:7}, the results validate that our method enhances the stability of soft prompts under various initialization strategies. The standard deviations of our method on target domain SST-2 and QNLI are 1.11 and 0.37 lower than those of the vanilla soft prompt tuning, and the performance is better compared with the vanilla soft prompt tuning.
\begin{table}[h]
\centering
\begin{tabular}{l|cc|cc}
\hline
              & \multicolumn{2}{c|}{SST-2}                                                           & \multicolumn{2}{c}{QNLI}                                                              \\ \hline
Methods       &  PT with both & Prompt Tuning & PT with both & Prompt Tuning \\ \hline
Random &                         \textbf{69.36}                     & 65.51                   & 50.31                & 49.60                     \\ 
Label                              & 68.67                       &  64.21                      & 50.84                      & 49.33                     \\ 
Vocab                        & \doubleunderline{66.88}                &  \doubleunderline{62.03}                & \textbf{51.20}                     & \textbf{50.45}                \\ 
Top-1k                     & 67.03              & 62.15             & \doubleunderline{50.05}             & 49.45             \\ 
Task                         &68.92                     & \textbf{67.29}                   &  50.10                &  \doubleunderline{48.04}                     \\\hline
Std.                         & 1.14                     & 2.25                   & 0.50                &  0.87                   \\\hline
\end{tabular}
\caption{Comparison of stability to soft prompt initialization. 
The best result across different templates is bold and the worst is double underline.
}
\label{tab:7}
\end{table}

\subsection{Extension to Decoder-only PLMs.}\label{D.3}
We explore the effectiveness of our approach on decoder-only PLMs. Keeping other experimental conditions unchanged, we replace the RoBERTa-Large with the GPT-2-Samll \citet{radford2019language} and perform the corresponding experiments.

\begin{table}[h]
\resizebox{\textwidth}{!}{
\begin{tabular}{ccccc|ccc}
\toprule
        &      & \multicolumn{3}{c|}{Sentiment}                                                           & \multicolumn{3}{c}{NLI}                                                              \\ \midrule
Paradigms & Methods       & S + M $\rightarrow$ C & C + M $\rightarrow$ S & S + C $\rightarrow$ M & Q+R$\rightarrow$ W & W+R$\rightarrow$ Q & Q+W$\rightarrow$ R \\ \midrule
\multirow{2}{*}{\makecell[c]{Prompt \\ Tuning}} &
Vanilla PT &   56.29$_{\text{1.03}}$                      & 67.57$_{\text{2.74}}$                       & 58.50$_{\text{2.31}}$                     & 42.64$_{\text{1.87}}$                   & 49.71 $_{\text{1.57}}$               & 49.48$_{\text{0.63}}$                     \\ 
&PT with both     & 57.85$_{\text{0.91}}$              & 68.52$_{\text{2.77}}$              & 59.73$_{\text{2.63}}$              & 42.79$_{\text{1.21}}$             & 50.27$_{\text{1.13}}$             & 51.10$_{\text{0.57}}$             \\
\bottomrule
\end{tabular}
}
\caption{Decoder-only PLM (GPT-2-Samll) backbone tests in accuracy.}
\label{tab:8}
\end{table}

The results in \tabref{tab:8} show that our method works well on the decoder-only PLMs backbone and successfully outperforms representative soft prompt tuning.

\subsection{Attention Visualization}
We show in \tabref{tab:9} the \textit{concentration strength} (CS) and \textit{concentration fluctuation} (CF) obtained in the last layer of RoBERTa-Large before and after soft prompt are optimized using our method. 
The results indicate that in the SST-2 target domain, our method not only significantly enhances the \textit{concentration strength} of soft prompt, but also effectively reduces the \textit{concentration fluctuation}, thereby achieving significant performance improvements in the SST-2 target domain. 
However, in the QNLI target domain, the \textit{concentration fluctuation} of soft prompt increases slightly, resulting in a limited improvement in accuracy. 
This suggests that \textit{concentration strength} and \textit{concentration fluctuation} jointly affect the generalization ability of prompts, which is consistent with observations from our pilot experiments in \secref{pilot}.
\begin{table}[!h]
\centering
\renewcommand\arraystretch{1.3}
\begin{tabular}{ccccc} 
\toprule
Target &  Method & CS &  CF & ACC\%  \\
\midrule
\multicolumn{1}{c}{\multirow{2}{*}{\begin{tabular}[c]{@{}c@{}}SST-2\end{tabular}}} & Prompt Tuning           &0.523 &     0.062   &  65.51                                                                  \\
\multicolumn{1}{c}{}                     & PT with both            & 0.578                   & 0.054                  & 69.36        \\
 \midrule
\multicolumn{1}{c}{\multirow{2}{*}{QNLI}}                     &Prompt Tuning      & 0.505          & 0.061   & 49.83                                                               \\
\multicolumn{1}{c}{}                     &PT with both               & 0.537         & 0.062  & 50.31                                                                 \\
\bottomrule
\end{tabular}
\caption{Accuracy affected by \textit{concentration strength} (the larger the better) and \textit{concentration fluctuation} (the smaller the better) before and after using concentrative soft prompt optimization.}
\label{tab:9}
\end{table}

\begin{figure}[h]
  \centering
  \includegraphics[width=0.29\textwidth]{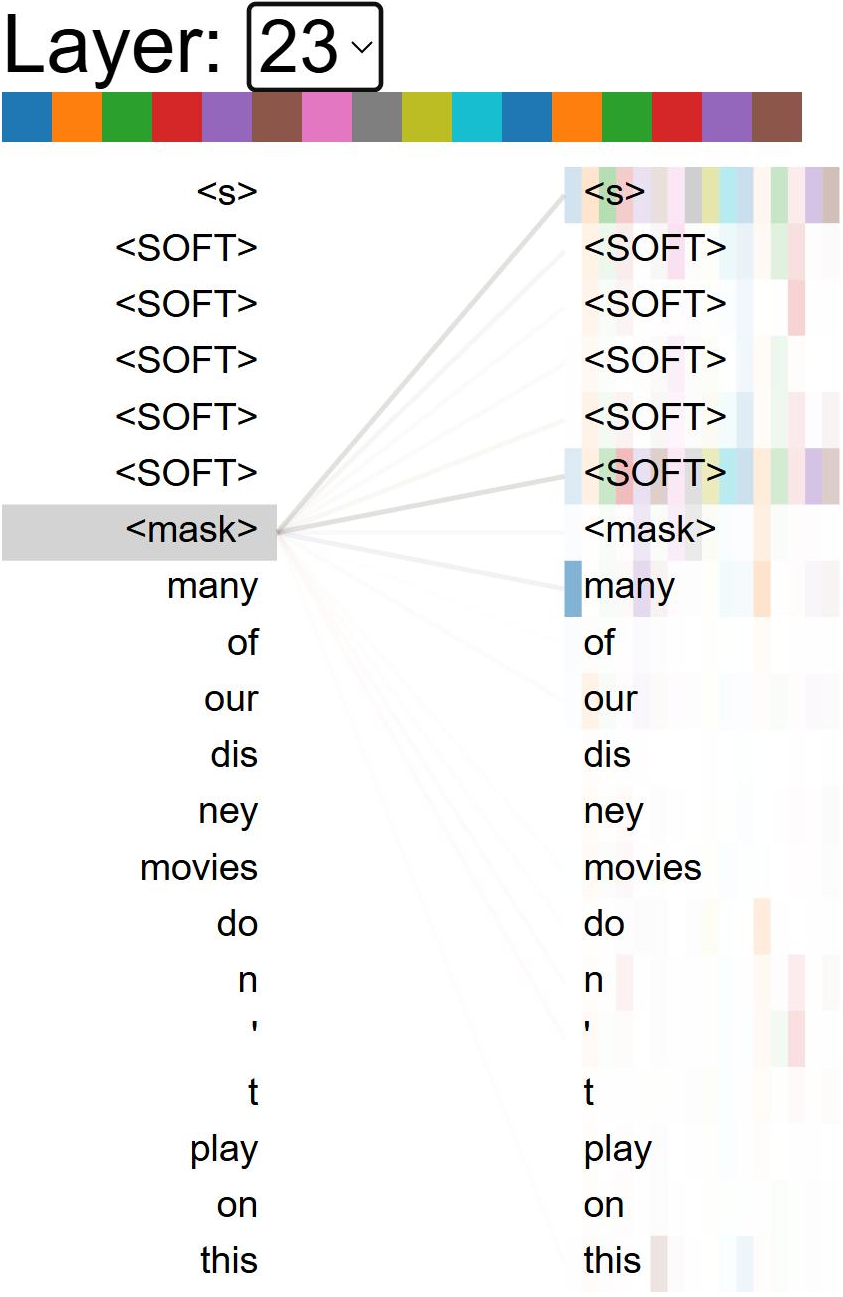}
  \hspace{20pt}
  \includegraphics[width=0.29\textwidth]{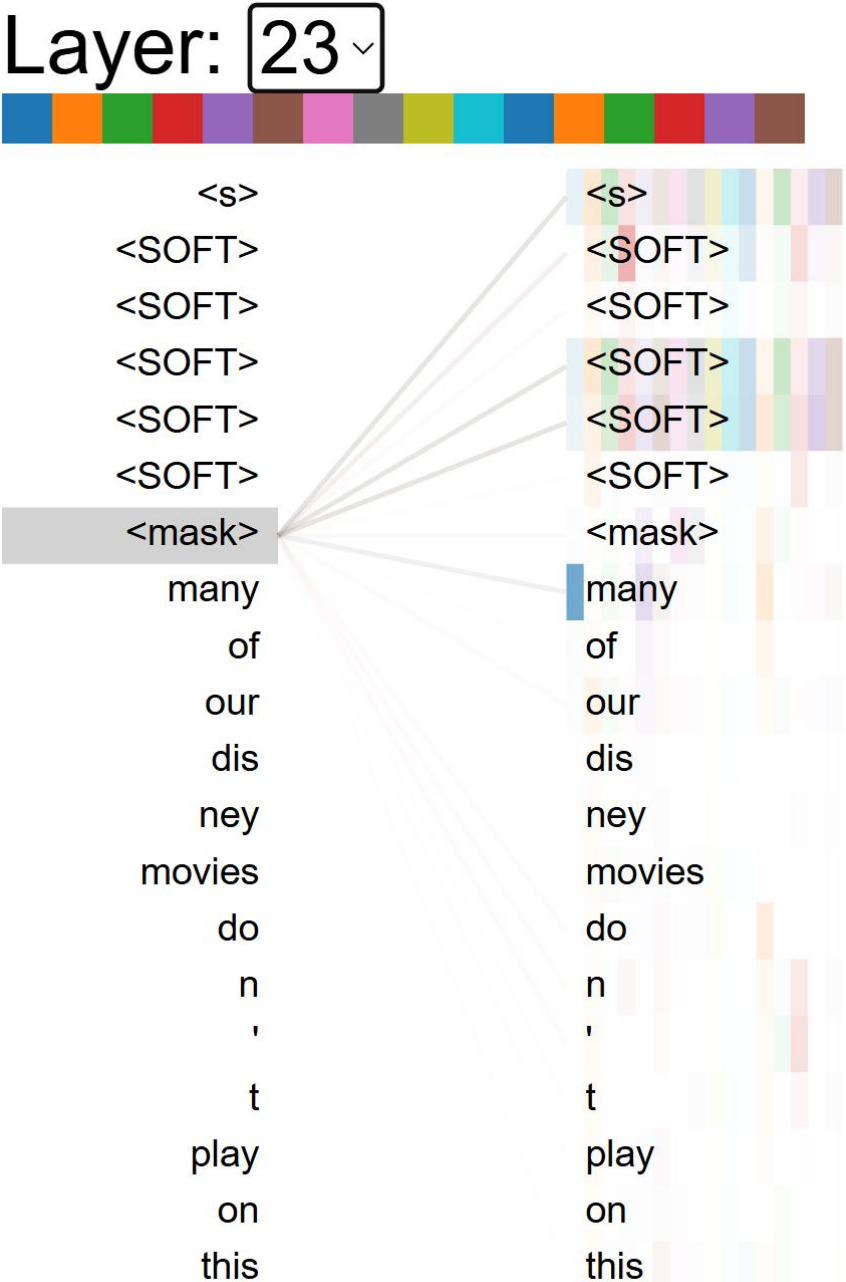}
  \caption{Case 1 for attention comparison visualization for soft prompt.}
  \label{sft1}
\end{figure}

\begin{figure}[htbp]
  \centering
  \includegraphics[width=0.3\textwidth]{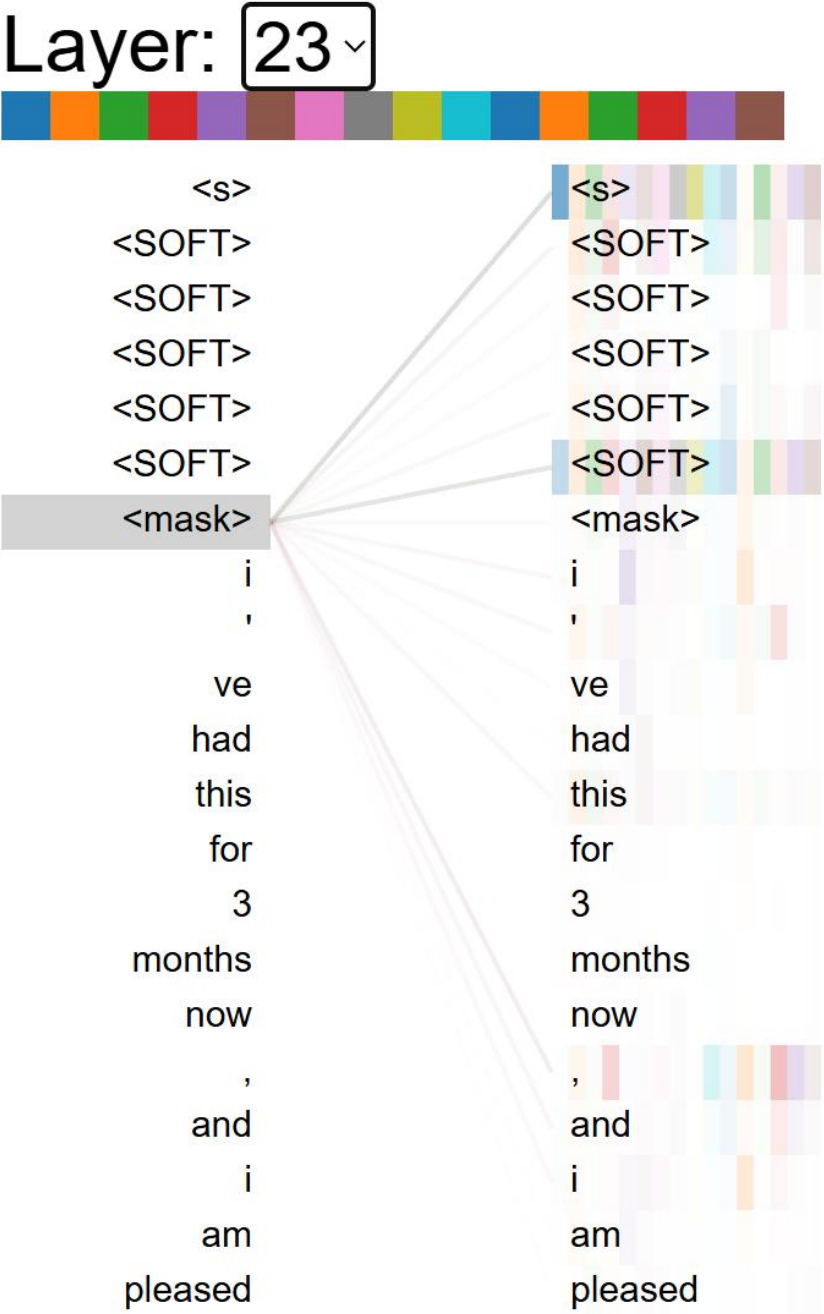}
  \hspace{20pt}
  \includegraphics[width=0.3\textwidth]{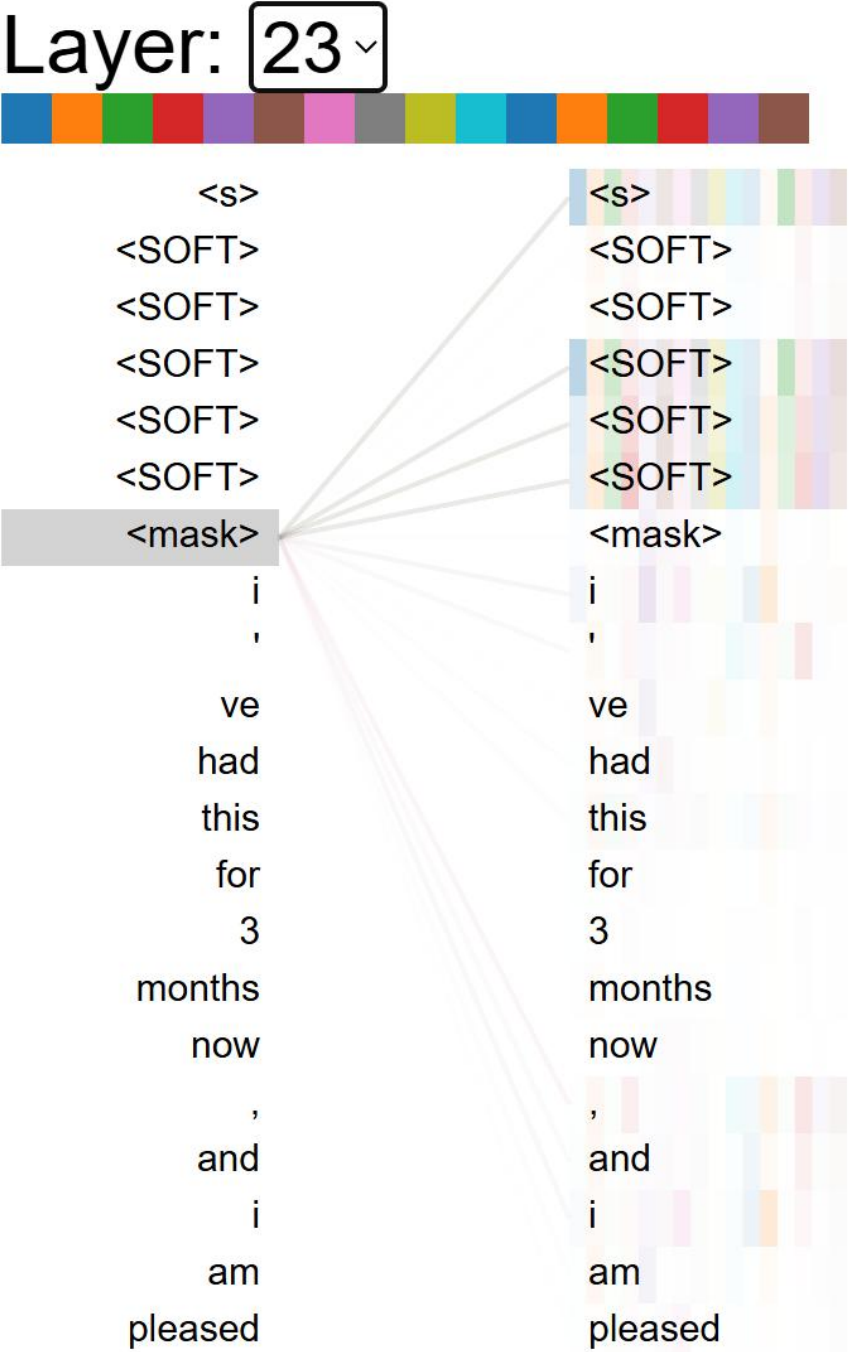}
  \caption{Case 2 for attention comparison visualization for soft prompt.}
  \label{sft2}
\end{figure}

\begin{figure}[htbp]
  \centering
  \includegraphics[width=0.3\textwidth]{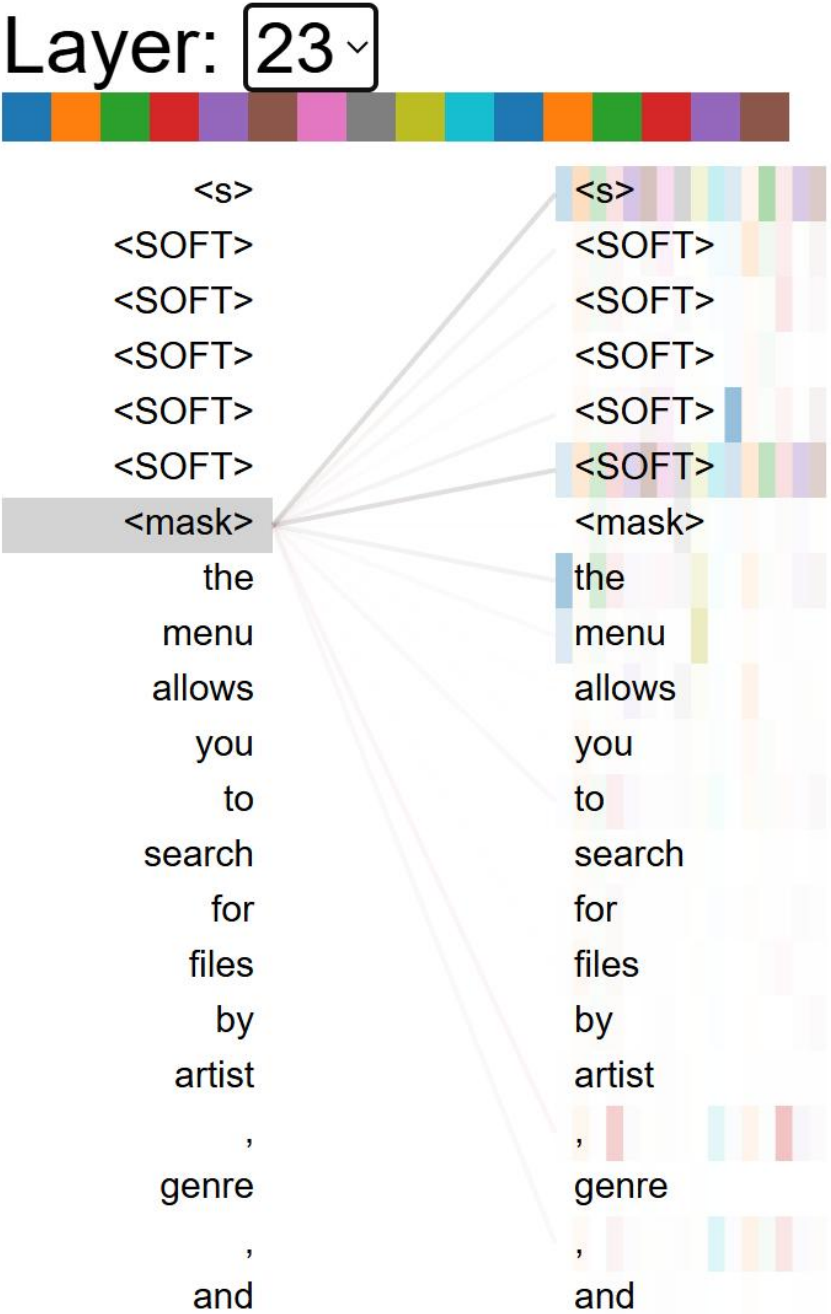}
  \hspace{20pt}
  \includegraphics[width=0.3\textwidth]{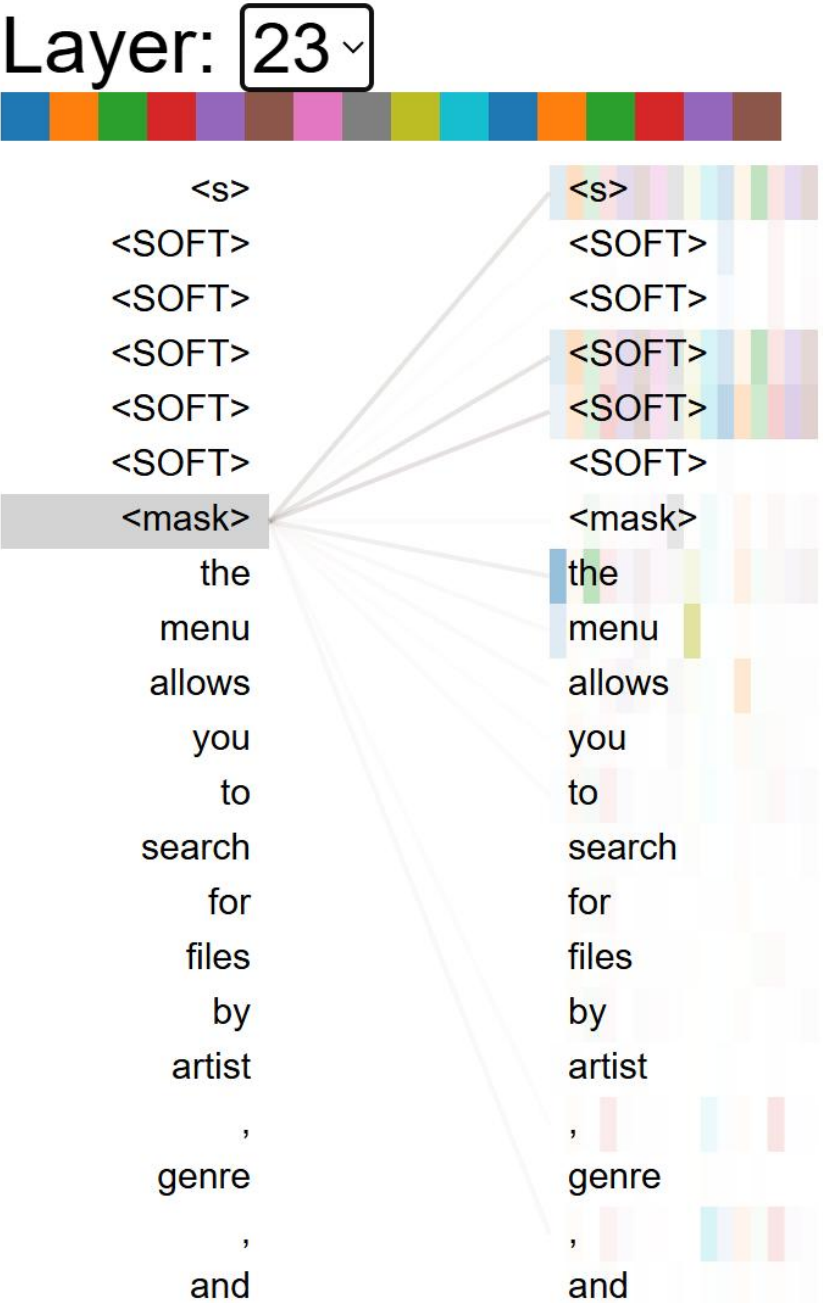}
  \caption{Case 3 for attention comparison visualization for soft prompt.}
  \label{sft3}
\end{figure}

\begin{figure}[htbp]
  \centering
  \includegraphics[width=0.3\textwidth]{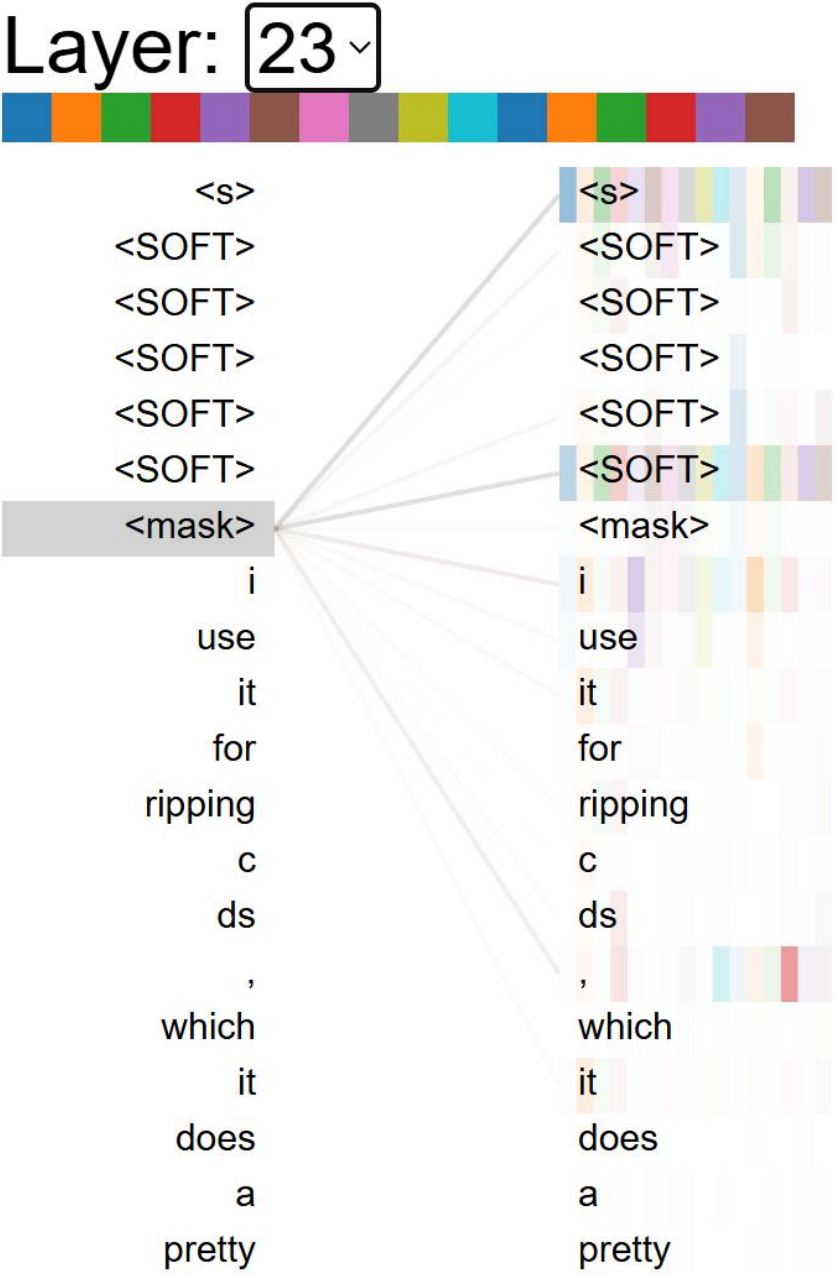}
  \hspace{20pt}
  \includegraphics[width=0.3\textwidth]{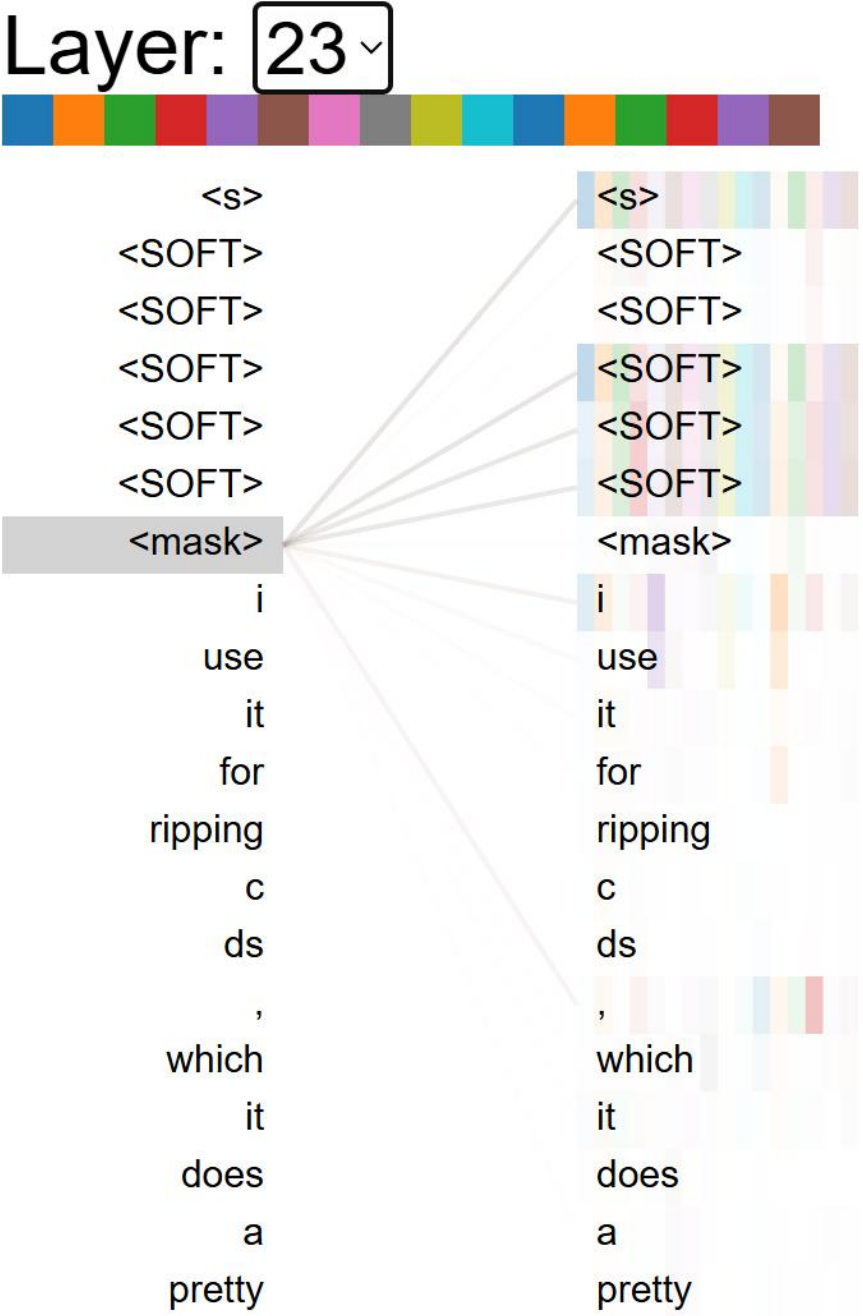}
  \caption{Case 4 for attention comparison visualization for soft prompt.}
  \label{sft4}
\end{figure}

\vspace{-0.6cm}
In addition, we also use bertviz \citet{vig-2019-multiscale} to visually display the continuous prompts before and after using concentrative soft prompt optimization.
In \figref{sft1} to \figref{sft4}, we show the attention distribution of vanilla soft prompt (left) and soft prompt trained with our method (right) on the same inputs at the last layer of the RoBERTa-Large model. 
It can be observed that concentrative soft prompt optimization improves the attention concentration and stability to soft prompts at predicted locations.

\section{Details for Concentrative Hard Prompt Optimization}\label{D}
\subsection{Optimization Process}
We define the prompt matching problem under MFDG setting as a multi-agent reinforcement learning (MARL) problem, as shown in Algorithm \ref{alg:algorithm2}.

\begin{algorithm}
\small
    \caption{Concentrative Hard Prompt Optimization} 
    \label{alg:algorithm2}
    \begin{algorithmic}[1]
        \State \textbf{Input:} Training set $\mathcal{D}_{\text{train}}$ of size $T$, testing set $\mathcal{D}_{\text{test}}$, fixed PLM parameterized by $\theta$, the prompt sets $\left\{\mathcal{Z}^{n}\right\}_{n=1}^{N}$ filtered by GCS \ref{GCS}, number of agents $N$.
        \begin{center}
            \textit{**** training the multi-agent RL model ****}
        \end{center}
        \State Initialize policy networks $\pi_{\omega_1}, \dots, \pi_{\omega_N}$ for each agent with parameters $\omega_1, \dots, \omega_N$ and $epoch \leftarrow 0$.
        \While{$epoch<epoch_{max}$}    
            \For{step $t$ \textbf{in} $[1,..., T]$}
                \For{each agent $n$ \textbf{in} $[1,..., N]$}
                    \State Get state $s_{t}^n \leftarrow \mathrm{PLM}(x_{t})$ for agent $n$.
                    \State Run policy network $\pi_{\omega_n}(a_{t}^n|s_{t}^n)$ to take an action $a_t^n$ to select a prompt $z_{t}^n$ from $\mathcal{Z}^{n}$.
                    \State Calculate reward for agent $n$, \ie Eq. \ref{reward}.
                    \State Add $(s_t^n, a_t^n, r_t^n)$ transition to agent $n$'s replay buffer.
                \EndFor
                \State  Update of parameters $\omega_1, \dots, \omega_N$ using the MAPPO algorithm \cite{yu2022surprising}.
            \EndFor
        \EndWhile
        \begin{center}
            \textit{**** testing phase begins ****}
        \end{center}
        \For{each input $(x_i, y_i)$ in $\mathcal{D}_{\text{train}}$}
            \For{each agent $n$ \textbf{in} $[1,..., N]$}
                \State Get state $s^n \leftarrow \mathrm{PLM}(x_i)$. 
                \State Run policy network $\pi_{\omega_n}(a^n|s^n)$ to take an action $a^n$ to select a prompt $z^n$ from $\mathcal{Z}^{n}$.
            \EndFor
            \State Get final prediction according to Eq. \ref{finalpred}.
        \EndFor
        \State \textbf{Output:} A trained policy network $\pi_{\omega_1}, \dots, \pi_{\omega_N}$, predictions for test inputs.
    \end{algorithmic}
\end{algorithm}

\subsection{Attention Visualization}
\begin{figure}[htbp]
  \centering
  \includegraphics[width=3cm,height=8.5cm]{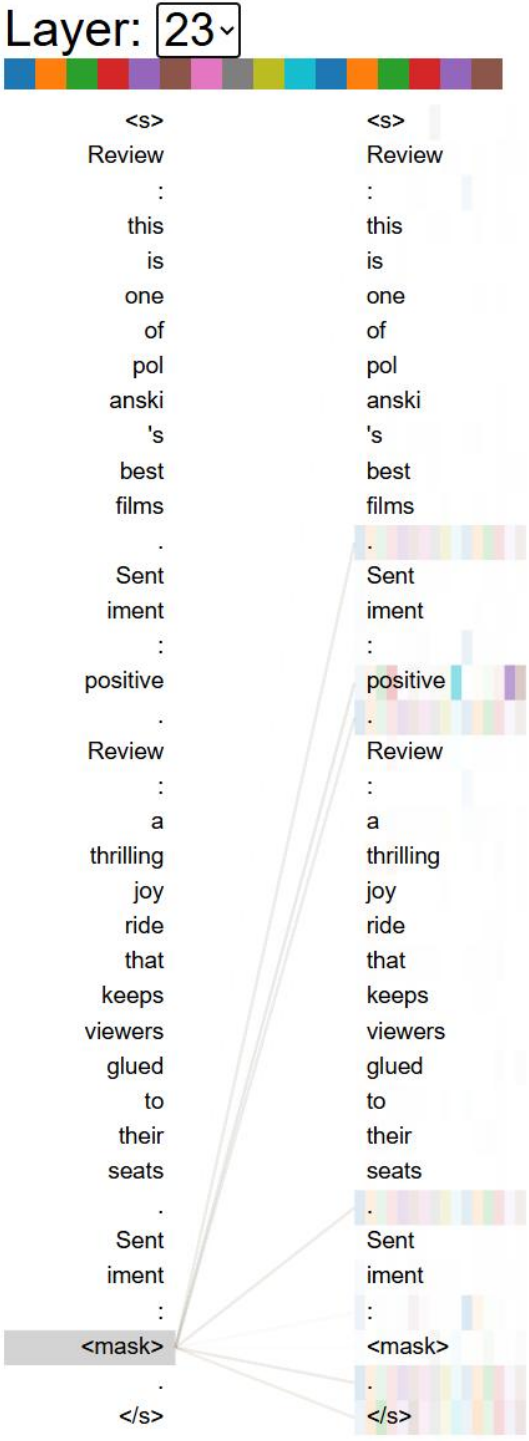}
  \hspace{20pt}
  \includegraphics[width=3cm,height=8.5cm]{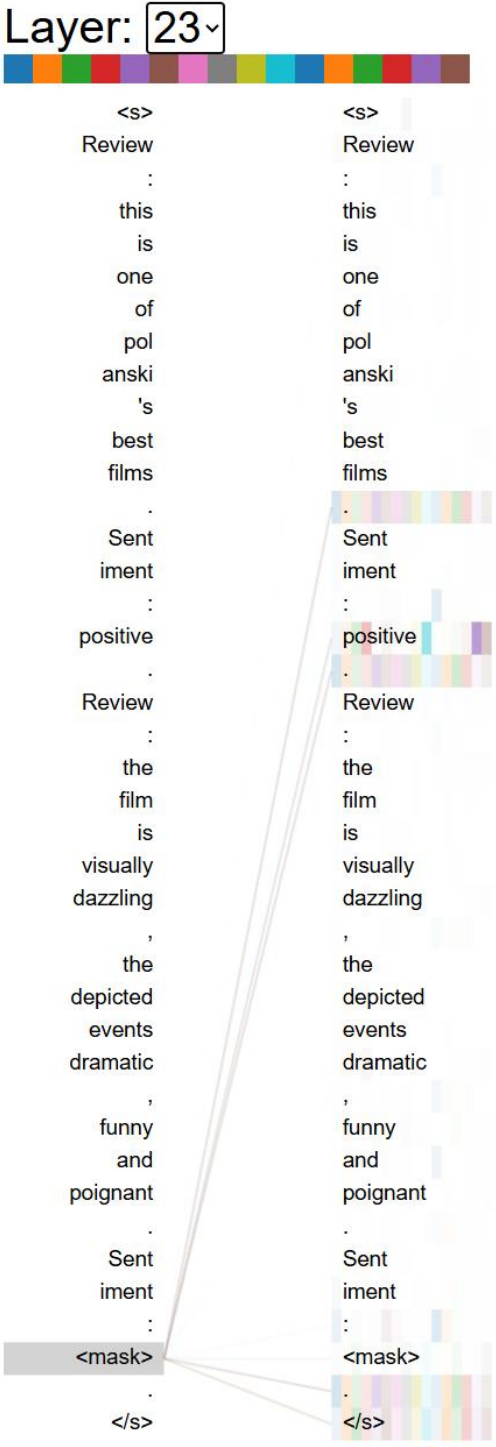}
  \hspace{20pt}
  \includegraphics[width=3cm,height=8.5cm]{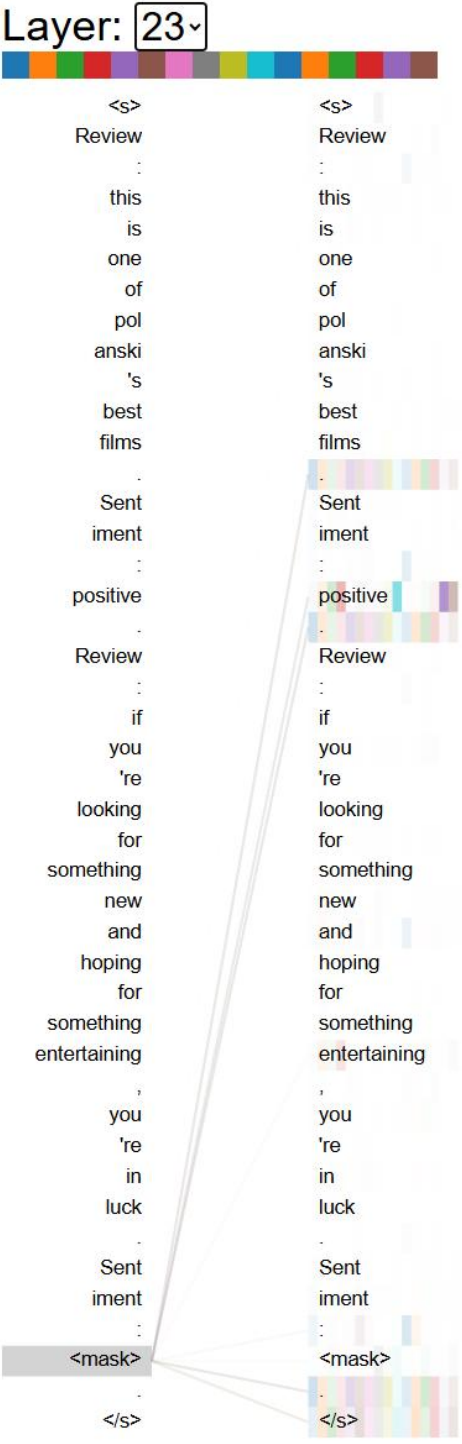}
  \caption{Case 1 for attention visualization of three filtered hard prompts.}
  \label{IC1}
\end{figure}

\begin{figure}[htbp]
  \centering
  \includegraphics[width=3cm,height=8.5cm]{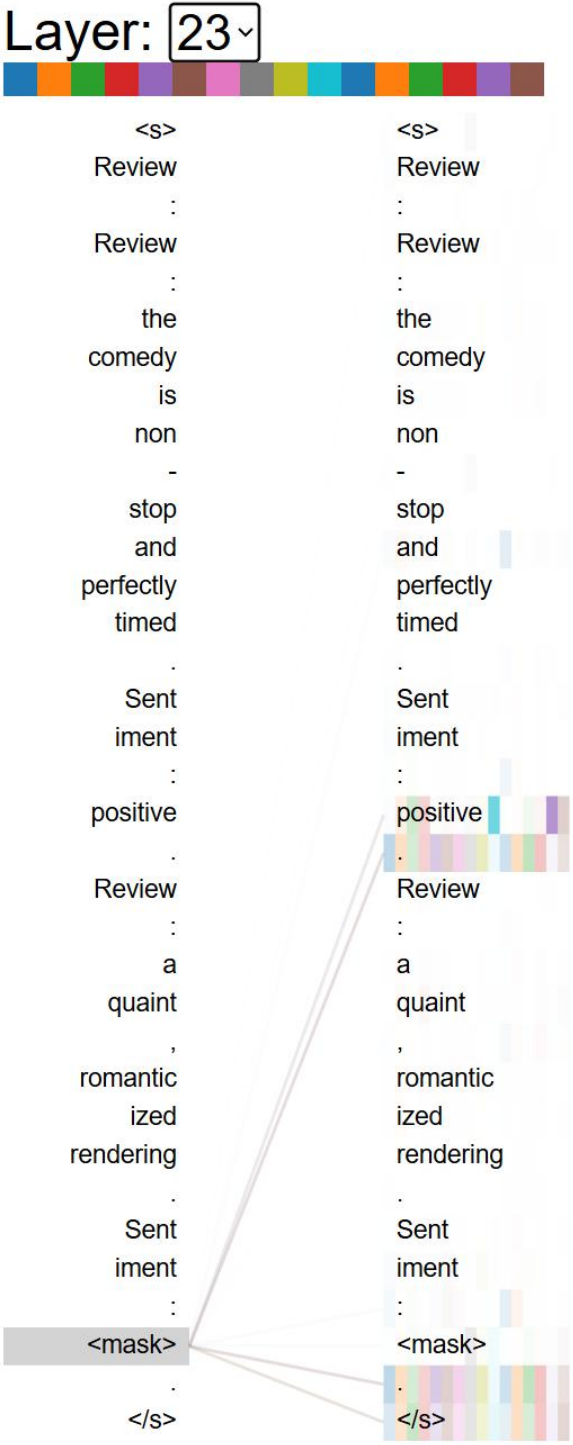}
  \hspace{10pt}
  \includegraphics[width=3cm,height=8.5cm]{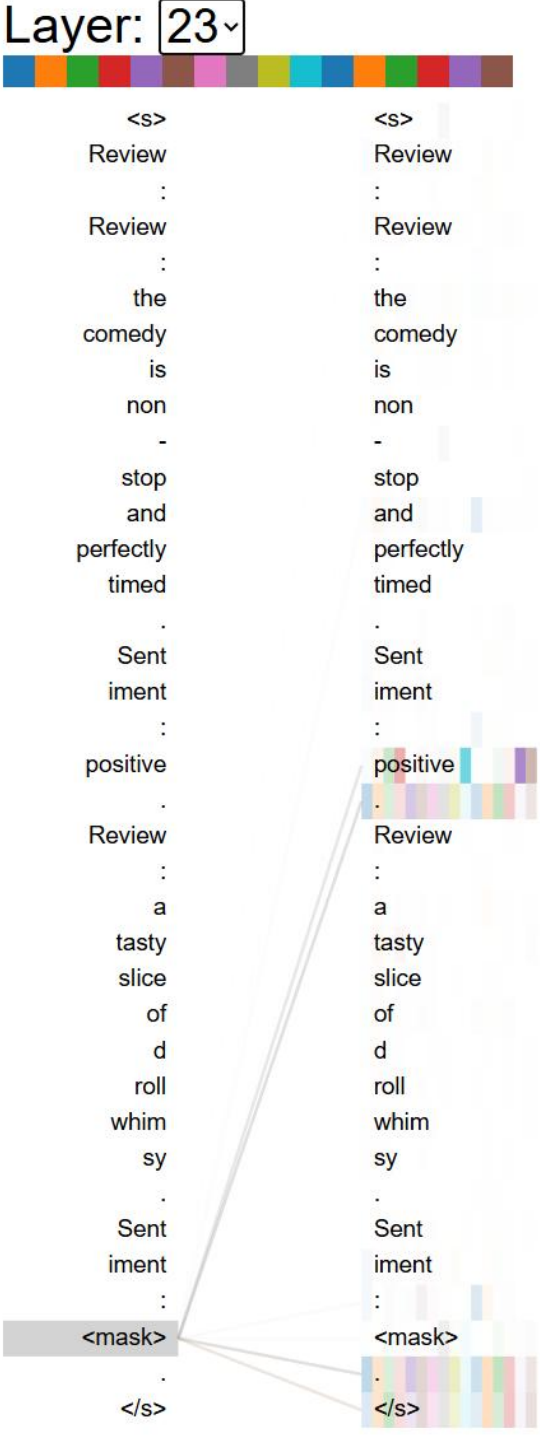}
  \hspace{10pt}
  \includegraphics[width=3cm,height=8.5cm]{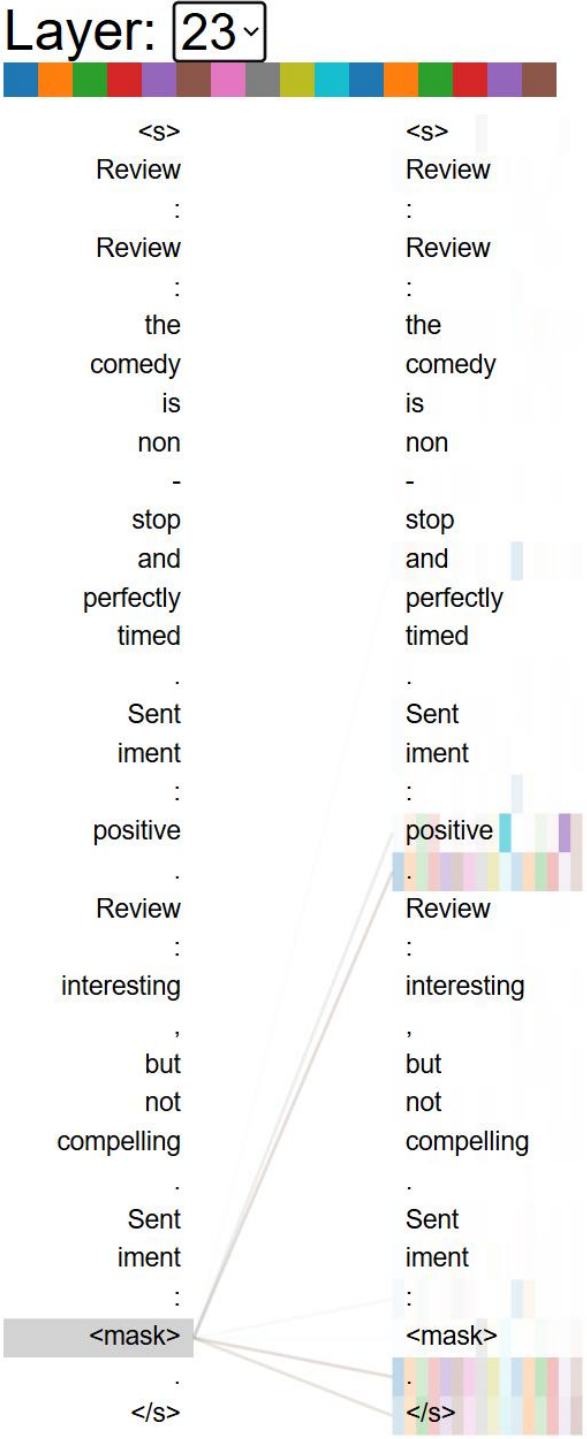}
  \caption{Case 2 for attention visualization of three filtered hard prompts.}
  \label{IC2}
\end{figure}

We utilize bertviz \citet{vig-2019-multiscale} to visualize the attention distribution in the final layer of the RoBERTa-Large model when processing different inputs with hard prompts filtered by GCS. As illustrated in \figref{IC1} to \figref{IC2}, the filtered hard prompts demonstrate high attention concentration and stability at the predicted positions.

\subsection{Stability to Hard Prompt Verbalizer Selection}
Prompt-based methods require mapping the probabilities generated by PLMs to the label space needed for downstream tasks.
Thus, the selection of a verbalizer significantly impacts the performance of PLMs \citet{liu2023gpt}.
Previous research \citet{xu2023making} explores the identification of appropriate verbalizers for these models. 
The experimental results in \tabref{Tab:verbalizers} show that our method achieves the highest accuracy under different verbalizers settings, which shows that our method can improve the robustness of existing methods for verbalizers selection.

\begin{table}[!h]
\centering
\small
\renewcommand\arraystretch{1.2}
\begin{tabular}{lcccc}
\toprule
\textbf{Verbalizer}         & \textbf{In
-Contex Demo}        & \textbf{IC with both}            & \textbf{$\textsc{DP}_2\textsc{O}$} & \textbf{$\textsc{DP}_2\textsc{O}$ with both}       \\
\midrule
\texttt{bad/good}            & 81.67$_{\text{1.12}}$          & \textbf{84.16$_{\text{1.02}}$}         & 89.36$_{\text{0.41}}$      & \textbf{91.73$_{\text{0.43}}$}    \\
\texttt{negative/positive} & 84.81$_{\text{1.39}}$          & \textbf{88.49$_{\text{1.55}}$}            & 90.75$_{\text{0.87}}$    & \textbf{92.87$_{\text{0.33}}$}        \\
\texttt{terrible/great}      & 83.32$_{\text{1.77}}$        & \textbf{87.72$_{\text{0.88}}$}            & 90.58$_{\text{0.62}}$   & \textbf{92.13$_{\text{0.91}}$}\\
\bottomrule
\end{tabular}
\caption{Analysis on stability to verbalizers.}
\label{Tab:verbalizers}
\end{table}

\subsection{Sensitivity to Number of Agents}
\begin{figure}[H]
\centering
\includegraphics[width=7cm]{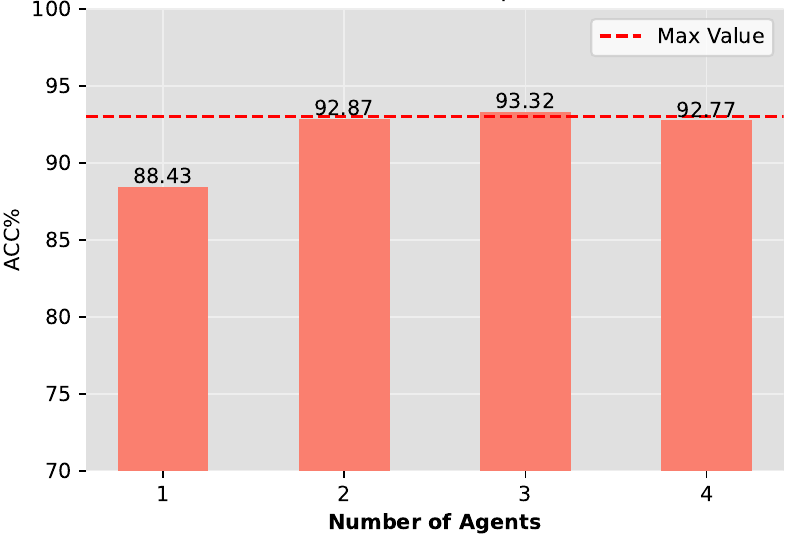} 
\caption{Performance for the model with different number of agents.}
\label{fig:num}
\end{figure}

We analyze the impact of the number of agents in a cue-matching framework. As shown in \figref{fig:num}, the experimental results reveal that when the number of agents is small, adding agents can significantly improve the classification accuracy of the target domain.
However, as the number of agents continues to increase, the accuracy gradually stabilizes.  
This suggests that as the number of prompts provided to the input gradually increases, the results of the ensemble decision will become more stable, and increasing or decreasing a prompt alone will have less impact on the overall performance of the prompt matching framework.
%or even begins to decline. 
%This suggests that too many agents may lead to over-integration of information or interference in the decision-making process, thereby weakening the overall performance of the system \citet{lee1998stability}.
%In addition,  redundant calculations between agents may also be one of the factors leading to the decrease in accuracy \citet{bai2024reducing}.

\section{Distribution of Concentration Strength on Larger Language Models}\label{E}
\begin{figure}[!ht]
\centering
\includegraphics[width=6.5cm]{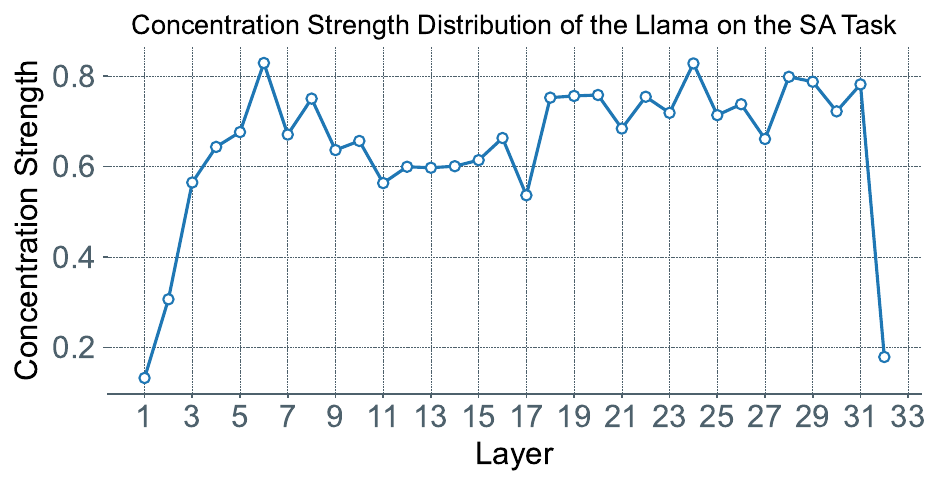} 
\includegraphics[width=6.5cm]{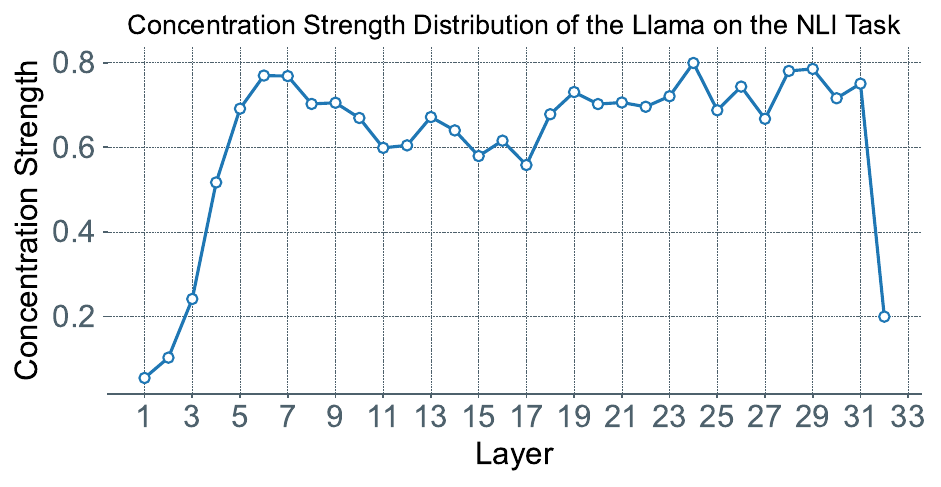}
\includegraphics[width=6.5cm]{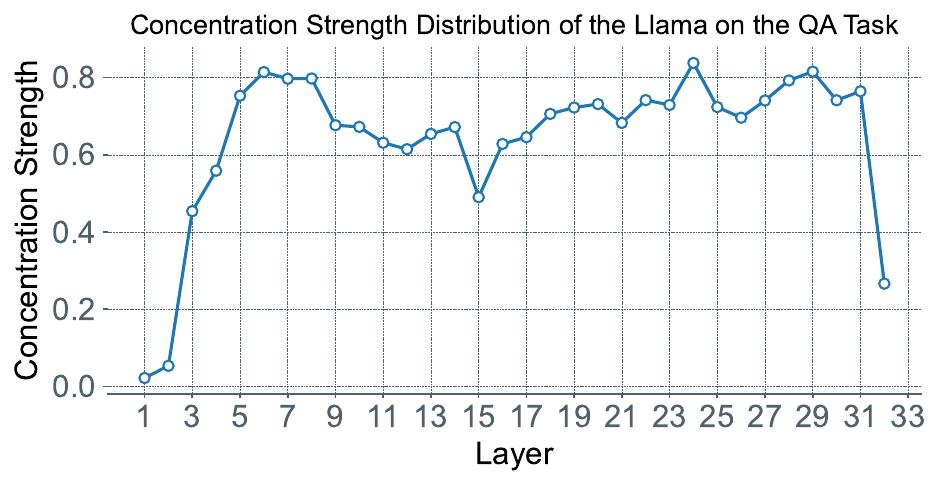}
\label{fig:istribution}
\caption{Concentration strength distribution of each layer of Llama in various tasks.}
\label{fig:LLM 1}
\end{figure}

As shown in \figref{fig:LLM 1} to \figref{fig:LLM 3}, we present the concentration strength distribution of three prominent open-source LLMs: Llama-2-7b-chat, Vicuna-7b-v1.5, and Alpaca-7b-wdiff. Our findings reveal that almost all three LLMs demonstrate higher concentration strength in deeper layers compared to shallower ones when processing prompts from different tasks. Moreover, larger models exhibit this concentration phenomenon earlier than smaller models.

\begin{figure}[!ht]
\centering
\includegraphics[width=6.5cm]{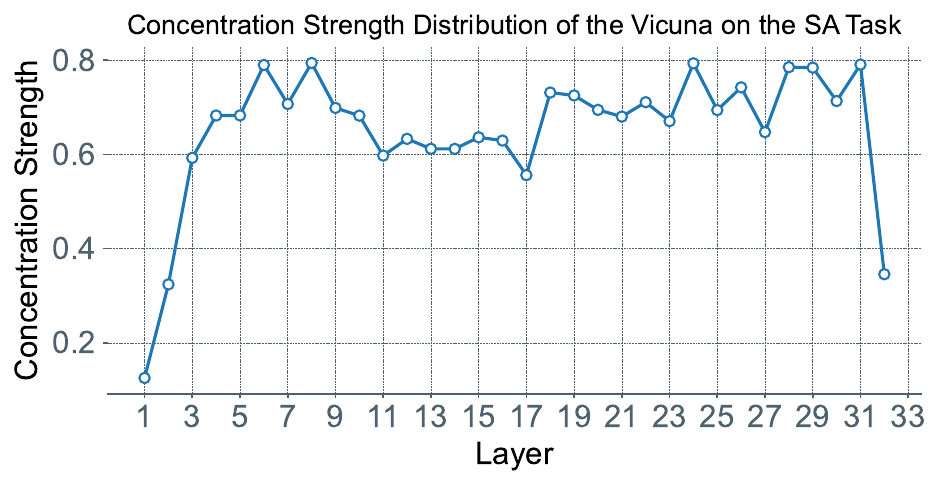} 
\includegraphics[width=6.5cm]{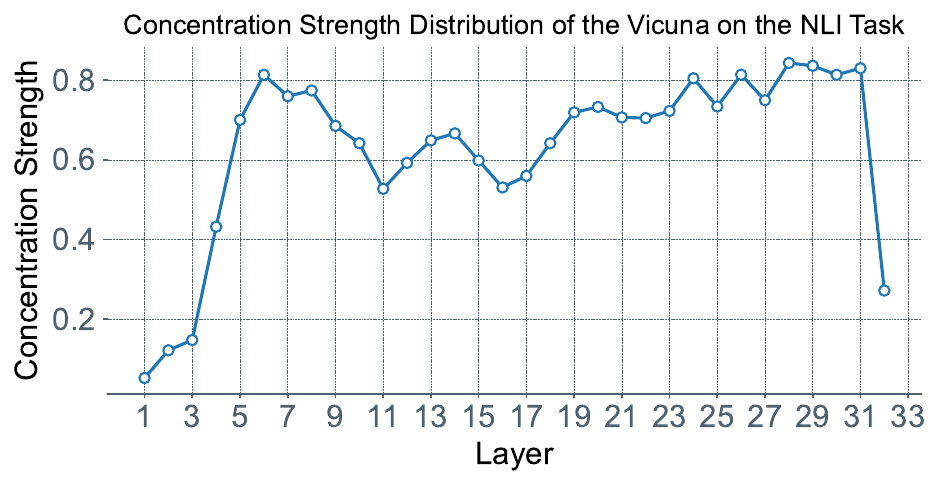}
\includegraphics[width=6.5cm]{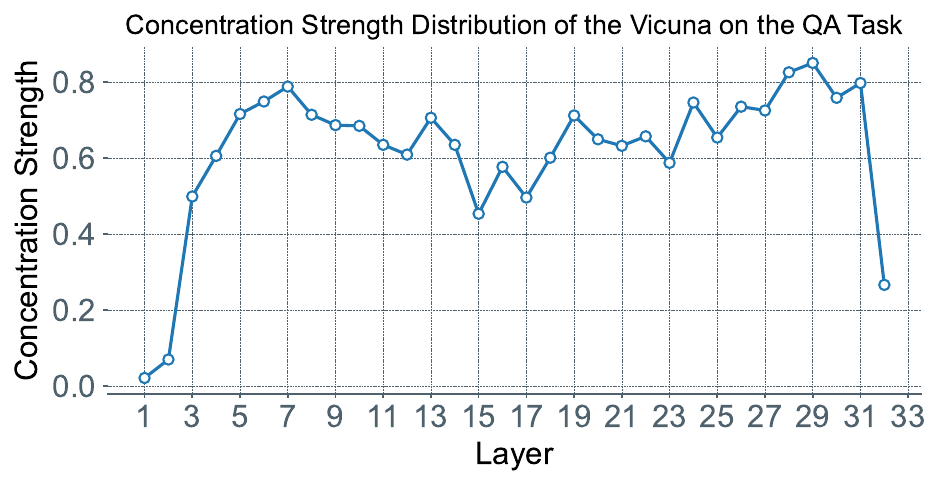}
\label{fig:LLM 2}
\caption{Concentration strength distribution of each layer of Vicuna in various tasks.}
\label{fig:istribution2}
\end{figure}
\begin{figure}[!ht]
\centering
\includegraphics[width=6.5cm]{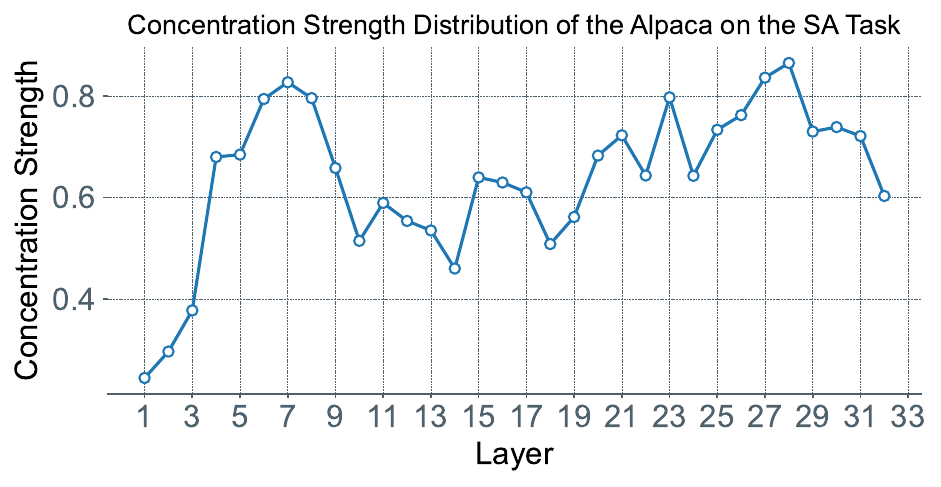} 
\includegraphics[width=6.5cm]{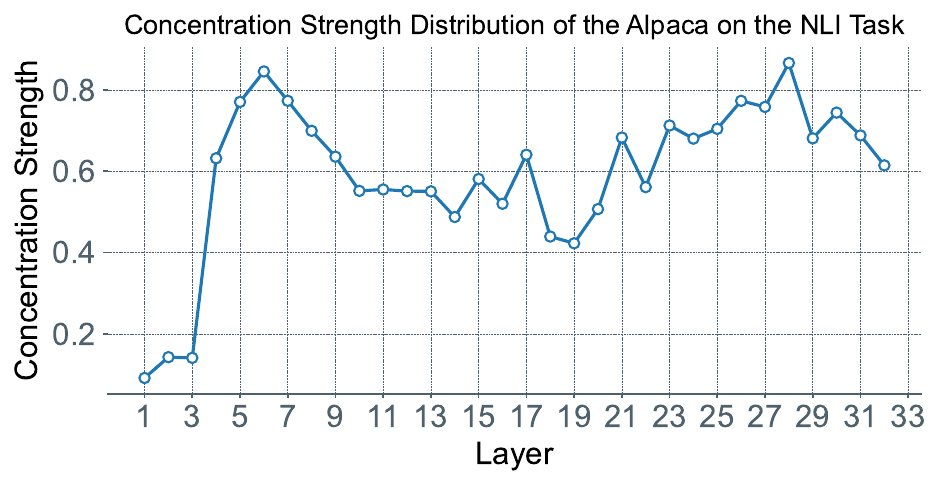}
\includegraphics[width=6.5cm]{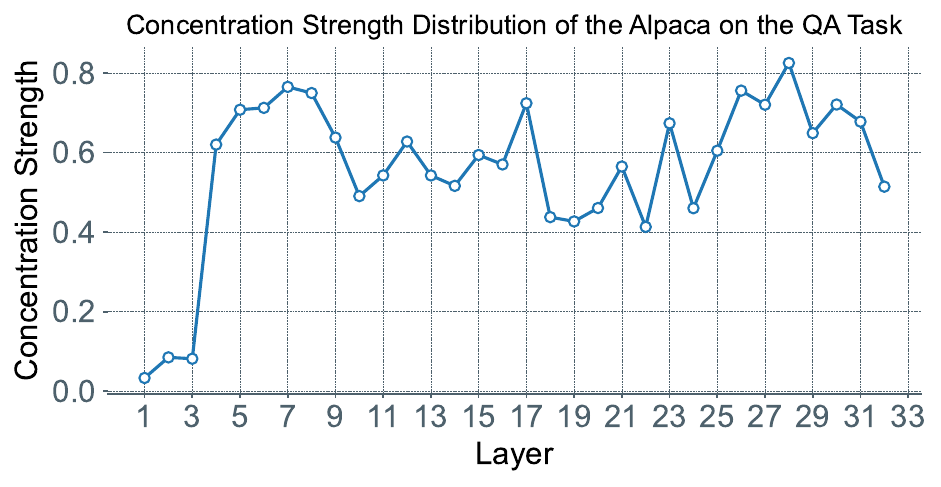}
\label{fig:LLM 3}
\caption{Concentration strength distribution of each layer of Alpaca in various tasks.}
\label{fig:LLM 3}
\end{figure}

\end{document}